\documentclass[preprint,5p,times,twocolumn,authoryear]{elsarticle}
\usepackage{amssymb}
\usepackage{amsmath}
\usepackage{multirow}
\usepackage{booktabs}
\usepackage{color}
\usepackage{lineno, hyperref}
\usepackage{bm}

\begin{document}

\begin{frontmatter}
\title{A Brain-to-Population Graph Learning Framework for Diagnosing Brain Disorders}

\author[label1]{Qianqian Liao}
\author[label1]{Wuque Cai}
\author[label1]{Hongze Sun}
\author[label1]{Dongze Liu}
\author[label1,label2]{Duo Chen}
\author[label1,label3]{Dezhong Yao\corref{cor1}}
\author[label1,label3]{Daqing Guo\corref{cor1}}
\address[label1]{Clinical Hospital of Chengdu Brain Science Institute, MOE Key Lab for NeuroInformation, China-Cuba Belt and Road Joint Laboratory on Neurotechnology and Brain-Apparatus Communication, School of Life Science and Technology, University of Electronic Science and Technology of China, Chengdu 611731, China.}
\address[label2]{School of Artificial Intelligence, Chongqing University of Education, Chongqing 400065, China.}
\address[label3]{Research Unit of NeuroInformation (2019RU035), Chinese Academy of Medical Sciences, Chengdu 611731, China.}
\cortext[cor1]{Corresponding authors: dyao@uestc.edu.cn (Dezhong Yao) and dqguo@uestc.edu.cn (Daqing Guo).}

\begin{abstract}
Recent developed graph-based methods for diagnosing brain disorders using functional connectivity highly rely on predefined brain atlases, but overlook the rich information embedded within atlases and the confounding effects of site and phenotype variability. To address these challenges, we propose a two-stage Brain-to-Population Graph Learning (B2P-GL) framework that integrates the semantic similarity of brain regions and condition-based population graph modeling. In the first stage, termed brain representation learning, we leverage brain atlas knowledge from GPT-4 to enrich the graph representation and refine the brain graph through an adaptive node reassignment graph attention network. In the second stage, termed population disorder diagnosis, phenotypic data is incorporated into population graph construction and feature fusion to mitigate confounding effects and enhance diagnosis performance. Experiments on the ABIDE I, ADHD-200, and Rest-meta-MDD datasets show that B2P-GL outperforms state-of-the-art methods in prediction accuracy while enhancing interpretability. Overall, our proposed framework offers a reliable and personalized approach to brain disorder diagnosis, advancing clinical applicability.

\end{abstract}
\end{frontmatter}

\begin{figure*}[!t]
\centering
\includegraphics[width=1\linewidth]{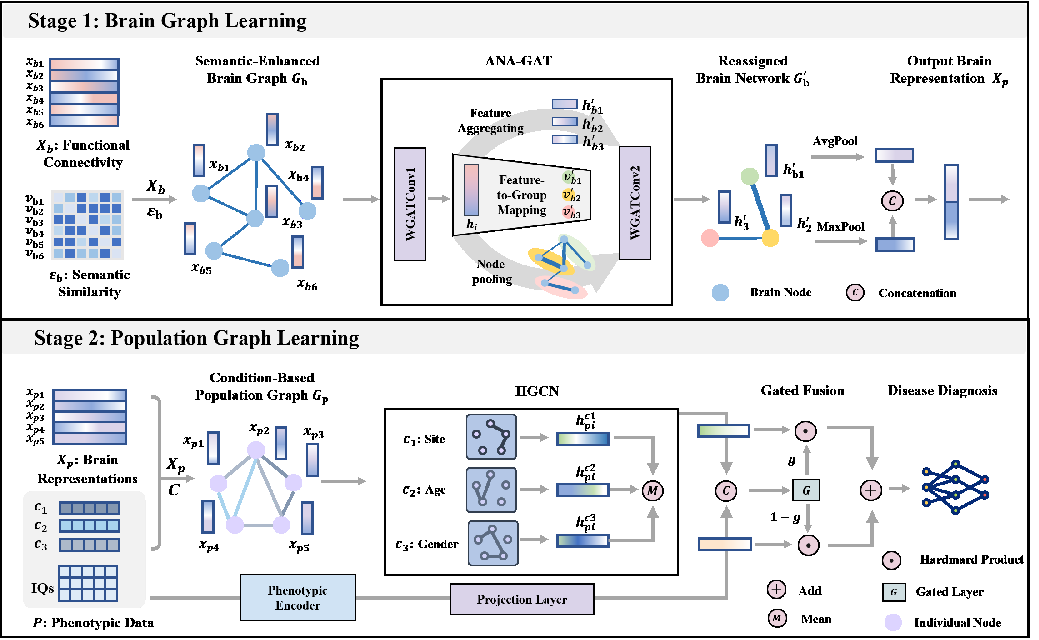}
\caption{The framework of our proposed Brain-to-Population Graph Learning Network. The upper section depicts the first stage, which constructs semantic-enhanced brain graphs and learns brain representations using the Adaptive Node Reassignment Graph Attention Network (ANR-GAT). The lower section illustrates the second stage, building a condition-based population graph and capturing meaningful patterns via a Heterogeneous Graph Convolutional Network (HGCN) and a gated fusion mechanism to integrate phenotypic data.}
\label{framework}
\end{figure*}

\section{Introduction}
Brain disorders—including neurodevelopmental conditions such as Autism Spectrum Disorder (ASD)\citep{Hodges2020Autism} and Attention-Deficit/Hyperactivity Disorder (ADHD) \citep{Hinshaw2018Attention}, as well as psychiatric disorders like Major Depressive Disorder (MDD)—significantly impair cognitive, emotional, and social functioning. These conditions not only limit individuals' potential but also place substantial burdens on families, communities, and healthcare systems \citep{Arora2018Neurodevelopmental}. Early and accurate identification and intervention are critical to mitigate these impacts and improve outcomes for affected individuals \citep{Finlay-Jones2019Very}.

Functional magnetic resonance imaging (fMRI) is a non-invasive technique for measuring brain activity through blood-oxygen-level-dependent (BOLD) signals \citep{logothetis2003underpinnings}, offering critical insights into complex neurological disorders \citep{park2013structural, arbabshirani2017single, tang2023comprehensive}. As a widely used metric for quantifying synchronized BOLD activity, functional connectivity (FC) assesses interregional communication by analyzing temporal correlations between brain regions partitioned using predefined atlases \citep{roland1994brain, logothetis2004interpreting}. However, existing approaches to FC analysis fail to utilize the rich medical information in these atlases, such as functional characteristics and structural organization of brain regions, which could improve the interpretation of fMRI data \citep{zhang2024novel}. Moreover, a major challenge in utilizing fMRI databases for neuroscience research lies in the presence of confounding effects, arising from site-related differences (e.g., MRI machine vendors) and population heterogeneity (e.g., age, gender) \citep{gordon2021three, geerligs2017challenges, masi2017overview}. These confounds \citep {rao2017predictive, brown2012adhd} may cause models to rely on spurious correlations rather than clinically meaningful measures \citep{ferrari2020dealing}.

Numerous computer-aided diagnosis methods have been developed based on FC analysis for diagnosing neurological disorders. Among these, graph neural networks (GNNs) have become a dominant approach owing to their powerful graph embedding capabilities. Building on this strength, recent works \citep{scarselli2008graph} and \citep{cui2022braingb} have demonstrated the effectiveness of GNNs in analyzing both brain networks and population networks. However, two main limitations exist in current brain network GNNs: (1) the underutilization of essential information encoded in brain atlases, and (2) the sole focus on individual FC embeddings, neglecting the inter-individual relationships both within and across groups \citep{wang2024multiview}. Although the population GNNs incorporate non-imaging data to model relationships between subjects, they often rely on linking individuals from the same sites or with similar demographics \citep{evgcn, lggnn}. Given that GNNs operate through neighborhood aggregation \citep{chen2024fairness}, this approach amplifies the biases inherent in FC data, where confounding factors such as site and demographic variations already mask disease-relevant patterns and may make it harder for the model to learn class-discriminative features \citep{kazi2019inceptiongcn}.

Moreover, despite the critical biological insights encoded in brain atlases, their potential remains underutilized due to the absence of methods for extracting interpretable and computationally efficient representations, coupled with the need for standardized and precise descriptions of brain regions. Recent advances in large language models (LLMs) have demonstrated promise in medical and neuroscience applications, offering potential solutions to bridge this gap \citep{achiam2023gpt}. These models can distill vast knowledge repositories into compact representations, serving as robust foundations for downstream tasks \citep{bzdok2024data}. Textual descriptions can additionally be encoded into high-dimensional vectors via embedding models, effectively capturing contextual relationships \citep{Chadha2021Distilled}. Therefore, LLMs have been successfully deployed to generate precise explanations of medical concepts \citep{yan2023robust} and compute semantic similarities between them \citep{peng2024mmgpl}. However, their direct application to information in the brain region remains largely unexplored.

To address these aforementioned challenges, we propose a novel two-stage Brain-to-Population Graph Learning (B2P-GL) framework. In the first stage, termed brain representation learning, we leverage GPT-4 and the embedding model to construct a semantic-enhanced brain graph by calculating semantic similarities between brain regions. The graph is refined using the Adaptive Node Reassignment Graph Attention Network (ANR-GAT), which captures latent node relationships while improving the interpretability of the brain network. In the second stage, termed population disorder diagnosis, we construct a condition-based heterogeneous graph by connecting nodes with high similarity in brain features but different categories within one condition. This structure enables a Heterogeneous Graph Convolutional Network (HGCN) to aggregate cross-condition messages while mitigating confounding effects. We further integrate phenotypic data via gated fusion to boost diagnostic accuracy. The proposed framework is validated on three large-scale, publicly available brain disorder datasets (ABIDE I, ADHD-200, and Rest-meta-MDD), evaluating both the ANR-GAT model and the B2P-GL framework to ensure adaptability across broader applications.

The main contributions of B2P-GL are as follows:

\begin{itemize}
    \item We propose the B2P-GL framework, which integrates a two-stage GNN to learn brain representations and population-level interactions, thereby improving both neuroscientific interpretability and diagnostic performance.
    \item We introduce semantic brain region embeddings and adaptive node reassignment for richer brain graph representations, and a condition-based population graph leveraging phenotypic data to reduce confounding and guide HGCN toward disorder-relevant patterns.
    \item We evaluate the proposed B2P-GL framework on three publicly available brain disorder datasets. Experimental results demonstrate that B2P-GL outperforms existing baselines, indicating its effectiveness and generalizability across diverse diagnostic tasks.

\end{itemize}

\section{Related Work}
\subsection{GNN-based Neurological Disorders Diagnosis}
Computer-aided diagnosis using resting-state functional MRI (rs-fMRI) data has gained increasing interest, with approaches ranging from traditional machine learning methods like support vector machines \citep{gallo2023functional} to deep learning techniques such as convolutional neural networks \citep{huang2023sd}, autoencoders \citep{yin2021diagnosis}, and transformers \citep{kan2022brain, bannadabhavi2023community}. Among these, GNN has emerged as a powerful framework for modeling brain networks \citep{braingnn} and population networks \citep{cao2021using}, due to their graph embedding capabilities and interpretability \citep{braingnn, qin2022using}. 

Researchers have explored multi-view FC across different atlases \citep{wang2024multiview}, multi-modal MRI data \citep{chen2022adversarial, zhu2022multimodal}, and dynamic FC \citep{campbell2023dyndepnet} to enrich FC representations in brain networks. However, these methods often require additional modalities or costly preprocessing. To improve brain network representation, advanced graph structures like hypergraphs \citep{ji2022fc, zuo2024prior} and heterogeneous graphs \citep{wen2024heterogeneous}, as well as learning methods such as metric learning \citep{ktena2018metric}, contrastive learning \citep{zhang2023gcl, shi2024contrastive, wang2025self}, self-supervised learning \citep{shi2025heterogeneous}, and adversarial learning \citep{chen2022adversarial, zuo2024prior}, have been applied to capture key patterns in FC data. In our brain representation learning, we enhance brain network representation by leveraging large language models (LLMs) to extract rich textual descriptions of brain regions, encoding their functional and structural characteristics. These embeddings provide additional semantic information, enriching the graph representations in brain network analysis. 

In population networks, neurological disorder prediction is treated as a node classification task. Node features are derived from the upper triangle of FC matrices \citep{shao2023heterogeneous} or extracted using methods like recursive feature elimination (RFE) \citep{yang2023diagnosis}. Alternatively, GNNs are employed to learn graph representations of FC for node feature embedding \citep{jiang2020hi, lggnn}. Non-imaging data, such as phenotypic features, are integrated by mapping them to edge weights  \citep{evgcn, lggnn, kazi2019inceptiongcn} or learning modality-shared and modality-specific representations to construct population networks \citep{zheng2022multi, chen2024deepasd}. While some recent studies have explored jointly modeling brain-level and population-level graphs to capture both within- and between-subject patterns \citep{pan2022mamf, li2024novel, zeng2025knowledge}, we use a two-stage approach to help disentangle subject-specific neural features from broader population-level variations.

\subsection{Neurological Disorders with Large Language Models}
Recent studies have explored LLMs in medical AI, particularly for tasks like medical image segmentation \citep{huang2024segment, zhang2023input}, using region-based prompts to identify regions of interest (ROIs). In neurological disorder diagnosis, MMGPL \citep{peng2024mmgpl} employs concept learning to filter irrelevant tokens by measuring semantic similarity to disease-related concepts. Moreover, fTSPL \citep{wang2024ftspl} leverages pre-trained vision-language models to enhance brain analysis, generating instance-level textual descriptions for each fMRI scan. While these methods demonstrate the effectiveness of LLMs in learning token-level or instance-level knowledge, the direct modeling of brain region semantics has received limited attention. To address this gap, we utilize LLMs to encode brain regions' textual descriptions, capturing their functional and structural characteristics. By computing semantic embeddings, we measure the similarity between brain regions based on their functional and structural profiles. This approach identifies connectivity patterns and coherence among regions with shared roles, offering new insights into brain network organization.

\section{Preliminary}
\subsection{Weighted Graph Attention in Brain Network}
\label{sec:brain}

Let $\mathcal{G}_b = (\mathcal{V}_b, \mathcal{E}_b, \bm{X}_b, \bm{W}_b)$ denote the brain network, where $\mathcal{V}_b$ denotes the set of brain nodes, $\mathcal{E}_b \subseteq \mathcal{V}_b \times \mathcal{V}_b$ is the set of edges between brain nodes, $\bm{X}_b \in \mathbb{R}^{|\mathcal{V}_b| \times F}$ is the node feature matrix with $F$ denoting the feature dimension. $\bm{W}_b \in \mathbb{R}^{|\mathcal{V}_b| \times |\mathcal{V}_b|}$ encodes the pairwise edge weights between nodes.

Graph Attention Network (GAT) employs a self-attention mechanism to dynamically learn the attention scores between neighboring brain nodes~\citep{ye2021sparse}. At layer $l$, the feature of brain node $i$ is denoted as $\bm{h}_{bi}^{(l)} \in \mathbb{R}^F$, and is first linearly transformed by a learnable weight matrix $\bm{W}^{(l)} \in \mathbb{R}^{F' \times F}$ to obtain: $\bm{z}_i^{(l)} = \bm{W}^{(l)} \bm{h}_{bi}^{(l)}$, where $\bm{W}^{(l)} \in \mathbb{R}^{F' \times F}$ is a learnable weight matrix, and $F'$ is the dimensionality of the transformed features. The attention score $e_{ij}$ between nodes $i$ and $j$ is computed as: $e_{ij} = \text{LeakyReLU} \left( \bm{a}^\top \left[ \bm{z}_i^{(l)} \| \bm{z}_j^{(l)} \right] \right)$, where $\bm{a} \in \mathbb{R}^{2F'}$ is a learnable parameter vector, and $\|$ denotes concatenation. Here, $\bm{z}_i^{(l)} \| \bm{z}_j^{(l)} \in \mathbb{R}^{2F'}$ is the concatenated feature vector of brain nodes $i$ and $j$.

To better incorporate the original edge weight information, we scale the attention score as:  $\tilde{e}_{ij} = w_{ij} \cdot e_{ij}$, where $w_{ij} \in \mathbb{R}$ represents the original normalized relation weight. We refer to this approach as the weighted graph attention network (WGAT) in later sections for brain representation learning. Finally, the updated feature for brain node $i$ at layer $l+1$, $\bm{h}_{bi}^{(l+1)} \in \mathbb{R}^{F'}$is computed as:
\begin{equation}
    \bm{h}_{bi}^{(l+1)} = \sigma \left( \sum_{j \in \mathcal{N}(i)} \frac{\exp(\tilde{e}_{ij})}{\sum_{k \in \mathcal{N}_b(i)} \exp(\tilde{e}_{ik})} \bm{z}_j^{(l)} \right)
\end{equation}
where the sum is taken over the brain node neighbors $\mathcal{N}_b(i)$, and $\sigma$ is a non-linear activation function. The output $\bm{h}_{bi}^{(l+1)} \in \mathbb{R}^{F'}$ represents the feature vector for node $i$ at layer $l+1$.

\subsection{Heterogeneous Graph Convolution in Population Network}
\label{sec:population}
The population graph is defined as $G_p = (\mathcal{V}_p, \mathcal{E}_p, \bm{X}_p, \mathcal{C})$, where $\mathcal{V}_p$ represents individual nodes, $\mathcal{E}_p$ represents population edges, $\bm{X}_p \in \mathbb{R}^{|\mathcal{V}_p| \times F_2}$ is the node feature matrix with $F_2$ denoting the input brain feature dimension and $\mathcal{C}$ is a set of conditions that define the relationships between nodes. A heterogeneous graph network (HGCN) \citep{schlichtkrull2018modeling, zhang2019heterogeneous} is used to propagate information across different relations. Each edge corresponds to a unique relationship under a specific condition $c \in \mathcal{C}$.

For individual node $i$, its feature representation $\bm{h}_{pi}^{c} \in \mathbb{R}^{F'_2}$ is updated by aggregating features from its condition-specific neighbors $\mathcal{N}_p(i)^c$, using a learnable weight matrix $W^c \in \mathbb{R}^{F'_2 \times F'_2}$ specific to condition $c$. HGCN ensures that the aggregation reflects the unique relational patterns within each condition. The final feature representation $\bm{h}_{pi} \in \mathbb{R}^{F'_2}$ is computed by mean aggregation across all conditions:

\begin{equation}
    \bm{h}_{pi} = \sigma\left( \frac{1}{|\mathcal{C}|} \sum_{c \in \mathcal{C}} \sum_{j \in \mathcal{N}_p^c(i)} \frac{1}{|\mathcal{N}_p^c(i)|} W^c \bm{h}_{pj} \right),
\end{equation}
where $\sigma$ is a non-linear activation function, and $\mathcal{N}_p^c(i)$ denotes the neighbors of individual node $v$ under condition $c$.

\section{Method}
The Brain-to-Population Graph Learning framework, as illustrated in Figure~\ref{framework}, uses a two-stage GNN approach to combine individual brain connectivity analysis with population insights. In the first stage, we model semantic-enhanced brain graphs using the ANR-GAT for brain representation learning (Section~\ref{sec:stage1}). In the second stage, we construct condition-based population graphs and leverage HGCN with gated fusion to incorporate phenotypic data for disease diagnosis (Section~\ref{sec:stage2}).

\subsection{Brain Representation Learning}
\label{sec:stage1}
As defined in Section ~\ref{sec:brain}, let $\mathcal{G}_\mathrm{b} = (\mathcal{V}_\mathrm{b}, \mathcal{E}_\mathrm{b}, \bm{X}_\mathrm{b}, \bm{W}_\mathrm{b})$ denote the brain graph, where brain nodes $\mathcal{V}_b$ correspond to parcellated regions of interest (ROIs) based on a brain atlas, and edges $\mathcal{E}_b$ capture the semantic relationships with weight $\bm{W}_b$, reflecting functional and structural connections between the atlas regions. This forms a semantic-enhanced brain graph. 
Each node $i$ has a feature vector $\bm{X}_{bi} \in \mathbb{R}^F$, with node features derived from the FC matrix, calculated using Pearson Correlation\citep{cohen2009pearson}, where each row represents the correlation coefficient between a node with all other nodes. 

\subsubsection{Semantic-Enhanced Brain Graph}

A common approach in healthcare involves consulting experts. However, recent advancements in instruction-following large language models, like GPT-4\citep{achiam2023gpt}, offer a promising alternative for addressing language-related challenges in the medical field\citep{nori2023capabilities}. By instructing GPT-4 with targeted prompts, we can generate lists of concepts linked to specific conditions, reducing annotation costs and utilizing large language models to tap into extensive medical knowledge\citep{peng2024mmgpl}. 

\begin{figure}[h]
    \centering
    \includegraphics[width=1\linewidth]{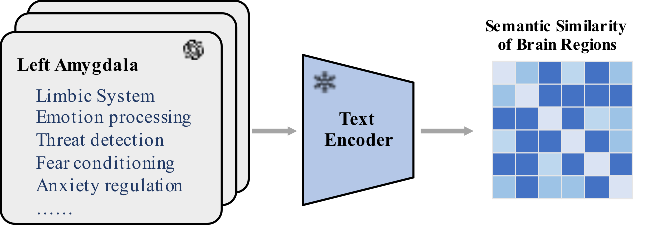}
    \caption{Illustration of semantic similarity generation for brain regions. Functional and structural descriptions (e.g., for the left amygdala) are encoded using a text encoder to produce embeddings. These embeddings are then used to compute semantic similarity between regions.}
    \label{fig:semantic}
\end{figure}

In this work, we leverage LLM to derive semantic embeddings of brain regions, capturing both their functional roles and structural properties.  For each brain region, prompts are constructed to elicit descriptive text capturing its core neurobiological attributes. These responses are then tokenized and embedded using the text encoder. An illustration of this process is provided in Figure~\ref{fig:semantic}. To quantify the semantic relationships between regions, we compute pairwise cosine similarity between their embeddings: $\text{Sim}(u, v) = \frac{\mathbf{s}_u \cdot \mathbf{s}_v}{|\mathbf{s}_u| |\mathbf{s}_v|}$, where $\mathbf{s}_u$ and $\mathbf{s}_v$, where $\mathbf{s}_u$ and $\mathbf{s}_v$ denote the semantic embeddings of brain regions $u$ and $v$, respectively. Based on these similarity scores, we construct edges in the brain graph, using a thresholded similarity score as the edge weight. This incorporation of semantic similarity enables the graph structure to more faithfully reflect the functional and structural relevance between regions.

The semantic similarities serve as a complementary source of information for FC, providing prior knowledge about potential anatomical and functional associations between regions. By integrating this knowledge-driven prior with data-driven FC features, the model is guided to interpret connectivity patterns more effectively and learns meaningful and generalizable representations beyond FC data, which is particularly beneficial in data-scarce or heterogeneous clinical populations.

\begin{figure}[h]
    \centering
    \includegraphics[width=1\linewidth]{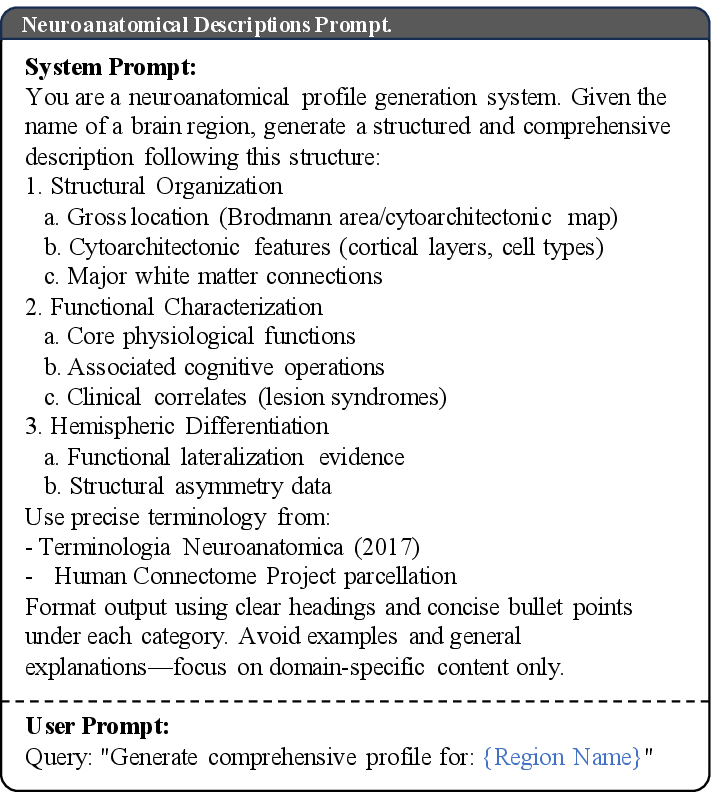}
    \caption{Prompt for generating neuroanatomical descriptions.}
    \label{fig:prompt}
\end{figure}

To ensure consistency and reproducibility in concept generation, we design a structured prompt format grounded in established neuroscience standards, such as the Terminologia Neuroanatomica and the Human Connectome Project. The prompt schema that guides the initial neuroanatomical descriptions is illustrated in Figure~\ref{fig:prompt}.

\begin{figure}[h]
    \centering
    \includegraphics[width=1\linewidth]{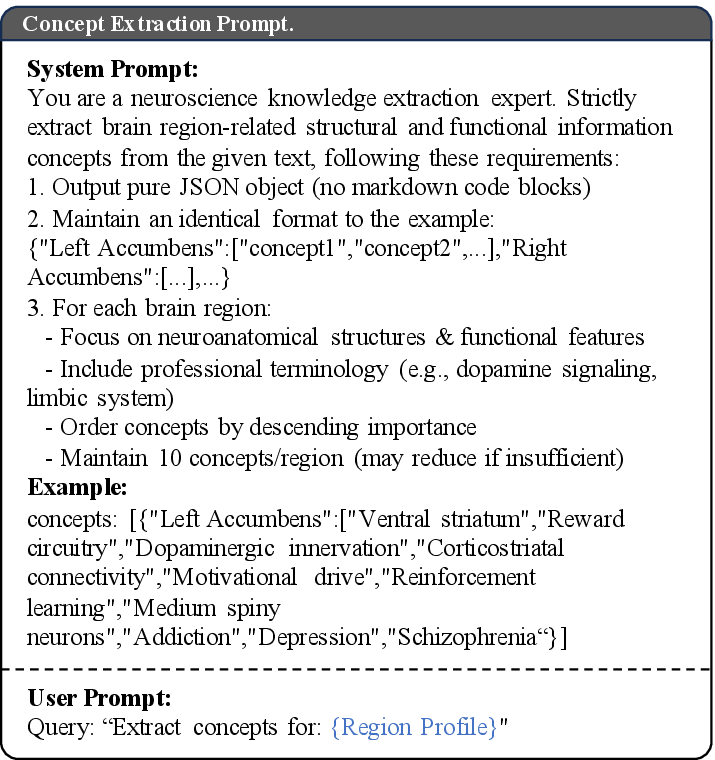}
    \caption{Prompt for structured extraction of neuroanatomical and functional concepts from text.}
    \label{fig:prompt2}
\end{figure}

To further enhance the granularity and generalizability of the extracted features, we introduce a secondary prompt tailored to extract hierarchical, ontology-driven concepts for each brain region. This prompt guides the model to produce fine-grained annotations that align with known structural-functional hierarchies. An overview of this concept extraction process is shown in Figure~\ref{fig:prompt2}. All generated descriptions were manually verified and cross-checked against standard references to ensure terminological consistency

To further enhance the granularity and generalizability of the extracted features, we introduce a secondary, ontology-driven prompt tailored to elicit hierarchical concepts for each brain region. This prompt guides the model to produce fine-grained annotations aligned with known structural–functional hierarchies, as shown in Figure~\ref{fig:prompt2}. The generated descriptions are manually verified to ensure terminological consistency.

In the ANR-GAT, meaningful features are extracted from neighboring nodes using WGAT, while feature-to-group mapping captures latent relationships between nodes to reassign the graph. Finally, brain representations are summarized using max and mean pooling across all nodes.

\subsubsection{Feature-to-Group Mapping}

As part of the ANR-GAT model, the node feature matrix $\bm{H}_b \in \mathbb{R}^{|\mathcal{V}_b| \times F'}$ is obtained after the WGAT layer. To reassign nodes into new groups based on their feature representations, we introduce a feature-to-group mapping mechanism that projects these features into a shared group space. The group assignment matrix $\bm{M} \in \mathbb{R}^{|\mathcal{V}_b| \times N_g}$, where $N_g$ is the number of groups, is computed by applying a learnable reassignment weight matrix $\bm{W}_r \in \mathbb{R}^{F' \times N_g}$ followed by a softmax operation: $\bm{M}=\text{Softmax}(\bm{H}_b \bm{W}_r)$, where the softmax is applied row-wise to ensure that the group assignment probabilities for each node sum to one.

For each node $i \in \mathcal{V}$, the group assignment $\text{group}_i$ is determined by selecting the group with the highest probability from $\bm{M}$ along the group dimension: $\text{group}_i = \arg\max_j M_{ij}, \quad \forall j \in \{1, 2, \dots, N_g\}.$ With this hard assignment mechanism, the actual group structure is adaptively learned from the data, and the effective number of groups may be smaller than the predefined upper bound $N_g$. As illustrated in Figure~\ref{framework}, this mapping is a key component of the ANR-GAT model, designed for brain graph representation learning. Node features are aggregated within each group, combining information from nodes reassigned to the same group. The edges are also remapped between groups, guided by the original graph connectivity and the reassigned groups. This process simplifies the graph structure while preserving key relationships, enabling the generation of new node representations that are passed to subsequent layers of WGAT for further processing. As an intermediate step bridging the WGAT layers, the feature-to-group mapping facilitates hierarchical feature learning and graph structure simplification.

\subsection{Population Disease Diagnosis}
\label{sec:stage2}
As defined in Section ~\ref{sec:population}, let $G_p = (\mathcal{V}_p, \mathcal{E}_p, \bm{X}_p, \mathcal{C})$, denote the condition-based population graph, where individual node features $\bm{X}_p$ represent brain representations from the first stage, and edges $\mathcal{E}_p$ are constructed based on the condition set $\mathcal{C}$ with node feature similarities to reduce biases while highlighting disease-specific patterns. We employ HGCN to capture key features within the population and use gated fusion to incorporate phenotypic features for enhanced diagnosis.

\subsubsection{Condition-Based Population Graph}
Given the heterogeneous nature of biomedical data, non-imaging variables or confounds \citep{rao2017predictive} can influence imaging data but may not be clinically relevant \citep{brown2012adhd}. These confounds introduce ambiguity in identifying the true predictive sources, potentially leading models to rely on confounds instead of the target measure \citep{ferrari2020dealing}. Unlike traditional approaches that directly connect nodes within the same category of a condition, our method leverages cross-category feature similarities to mitigate phenotype or site-related biases. This guides the model in learning intrinsic, disease-specific patterns rather than artifacts introduced by external conditions. To quantify the similarity of node features, we compute a scaled adjacency matrix $A$, derived from the pairwise feature distances and a Gaussian kernel: $A_{ij} = \exp \left( - \frac{\| \bm{h}_{pi} - \bm{h}_{pj} \|^2}{2 \sigma^2} \right)$, where $\bm{h}_{pi}$ and $\bm{h}_{pj}$ are the feature vectors of nodes $i$ and $j$, and $\sigma$ is a scaling parameter controlling the kernel width. 
Let $A^c$ denote the adjacency matrix for distinct categories within condition $c$, the entry $a_{ij}$ is defined as: $a_{ij}^c = \mathbb{I}(c_i \neq c_j)$, where $\mathbb{I}$ is the indicator function, equal to 1 if $c_i \neq c_j$ and 0 otherwise. Based on $A^c$, we select the top-$k$ most similar nodes to construct the overall condition-related edge index, minimizing biases for graph learning. 

The resulting population graph is trained using a heterogeneous GCN (HGCN) that is aware of condition types. HGCN assigns distinct transformation weights to different edge types, enabling the model to disentangle the effects of confounds and more effectively capture disease-relevant signals.

\subsubsection{Gated Fusion With Phenotypic Data}
The gated fusion mechanism adaptively combines graph-derived node features with encoded phenotypic features, ensuring that the final representation effectively integrates multimodal information while preserving the contributions of each source. Let $\bm{H}_p \in \mathbb{R}^{|\mathcal{V}_p| \times F_2}$ denote the graph node features and $\bm{P} \in \mathbb{R}^{|\mathcal{V}_p| \times D_p}$ represent the phenotypic features, where $|\mathcal{V}_p|$ is the number of individual nodes, $F_2$ is the dimension of the individual node features after the HGCN, and $D_p$ is the dimension of the phenotypic data. Phenotypic features are transformed into $\bm{P}' \in \mathbb{R}^{|\mathcal{V}_p| \times F_2}$ through a learnable network that maps $\bm{P}$ to the same dimensionality as $\bm{H_b}$. The concatenation of $\bm{H_b}$ and $\bm{P}'$ passes through a linear layer followed by a sigmoid activation to produce the gating coefficient $\bm{g}$:

\begin{equation}
    \bm{g} = \sigma(W_{\text{gate}} \cdot [\bm{H_b}, \bm{P}'] + b_{\text{gate}}),
\end{equation}
where $W_{\text{gate}}$ and $b_{\text{gate}}$ are the learnable weights and biases of the gating layer. The final representation is computed as a weighted combination of $\bm{H_b}$ and $\bm{P}'$, modulated by $\bm{g}$:
\begin{equation}
    \hat{\bm{H_b}} = \text{LayerNorm}(\text{ReLU}(\bm{g} \odot \bm{H_b} + (1 - \bm{g}) \odot \bm{P}')),
\end{equation}
where $\odot$ denotes element-wise multiplication. This mechanism enables dynamic integration of graph-derived and phenotypic information, enhancing the expressiveness and adaptability of the model.

\subsection{Two-Stage Optimization for B2P-GL Framework}
The proposed framework comprises two stages. The first stage is devoted to optimizing the representation of the brain graph, while the second stage aims to enhance the classification of disorders.

In brain graph learning, we formulate optimization as a joint problem that simultaneously learns node representations and enhances classification accuracy. Accordingly, the objective is to minimize a combined loss: $\mathcal{L}_1 = \mathcal{L}_{\mathrm{cls}} + \mathcal{L}_{\mathrm{sim}}$, where $\mathcal{L}_{\mathrm{cls}}$ is the classification loss and $\mathcal{L}_{\mathrm{sim}}$ is the similarity loss. This optimization enables the model to learn effective node embeddings while enhancing classification performance.

The classification loss xis computed using cross-entropy: $\mathcal{L}_{\mathrm{cls}} = -\frac{1}{N} \sum_{i=1}^{N} \sum_{c=1}^{2} \bm{y}_{i,c} \log \hat{\bm{y}}_{i,c}$, where $\bm{y}$ represents the ground truth labels and $\hat{\bm{y}}$ denotes the predictions.

\subsubsection{Similarity Loss}
The similarity loss aligns feature similarities with label similarities. Let $\bm{H} \in \mathbb{R}^{N \times d}$ represent node embeddings, with pairwise similarity defined as: $\bm{S}_{i,j} = \frac{\bm{h}_i \cdot \bm{h}_j}{\sqrt{d}}$. Row-normalizing $\bm{S}$ with the softmax function produces the normalized similarity matrix $\bm{S}'$. The loss is: 
\begin{equation}
    \label{equ:sim}
    \mathcal{L}_{\mathrm{sim}} = \frac{1}{N^2} \| \bm{S}' - \mathbb{I}[y_i = y_j] \|_F^2,
\end{equation}
where $\mathbb{I}[y_i = y_j]$ equals 1 if labels match and 0 otherwise. This loss encourages embeddings of nodes with the same label to be closer with different labels to be farther apart.

\subsubsection{Gradient of Similarity Matrix.}
The gradient of $\mathcal{L}_{\mathrm{sim}}$ adjusts embeddings based on similarity alignment. For node $i$:

\begin{equation}
\begin{split}
    \frac{\partial \mathcal{L}_{\mathrm{sim}}}{\partial \bm{h}_i} &= \frac{2}{N^2 \sqrt{d}} \sum_{j=1}^N \left( \bm{S}'_{i,j} - \mathbb{I}[y_i = y_j] \right) \bm{S}'_{i,j} \\
    &\quad \times \left( \bm{h}_j - \sum_{k=1}^N \bm{S}'_{i,k} \bm{h}_k \right).
\end{split}
\end{equation}

For the same label $\mathbb{I}[y_i = y_j] = 1$, the gradient pulls embeddings closer while the gradient pushes embeddings apart when $\mathbb{I}[y_i = y_j] = 0$.

\subsubsection{Compatibility of the Gaussian Kernel Similarity}
In the second stage, we construct the population graph by using the Gaussian kernel similarity measurement, which can be rewritten in terms of the dot product:
\begin{equation}
\begin{split}
    K(\bm{h}_i, \bm{h}_j) &= \exp\left(-\frac{2(1 - \bm{h}_i \cdot \bm{h}_j)}{2\sigma^2}\right) \\
    &= \exp\left(\frac{\bm{h}_i \cdot \bm{h}_j}{\sigma^2}\right) \cdot C,
\end{split}
\end{equation}
where $C = \exp(-1/\sigma^2)$ is a constant. This shows that maximizing $\bm{h}_i \cdot \bm{h}_j$ also maximizes the kernel similarity for same-class pairs.
The embeddings trained with $L_{\mathrm{sim}}$ inherently optimize dot-product similarities, which are directly compatible with the Gaussian kernel.

Considering the integration of the phenotypic, population graph learning extends the optimization from brain graph learning by independently aligning brain and phenotypic representations with label similarities while combining them with classification losses. The total loss function is $\mathcal{L}_2 = \mathcal{L}_{\mathrm{sim2}} + \mathcal{L}_{\mathrm{cls}},$  where  $\mathcal{L}_{\mathrm{sim2}}$ loss consists of a node representation similarity loss and a phenotypic fusion similarity loss, where the former ensures brain features capture disorder-relevant patterns, the latter aligns multimodal features with label similarities without enforcing consistency with the original brain features. This approach enables complementary learning of disease-related patterns from both modalities, enhancing predictive performance while preserving their unique contributions.

\section{Experiments}

In this section, we first provide a comprehensive overview of the experimental settings, including details on the publicly available preprocessed datasets, evaluation metrics, and implementation configurations. We then present extensive experiments to validate the effectiveness of the proposed framework and its individual components in two stages. The results demonstrate that B2P-GL achieves strong performance across three brain disorder datasets. Furthermore, ablation studies are conducted to examine the contribution of each module within the framework. Finally, we perform an interpretability analysis to explore clinically relevant patterns associated with different brain disorders.

\subsection{Datasets}

We evaluate our method on three widely used neuroimaging datasets related to neurodevelopmental and psychiatric disorders: Autism Brain Imaging Data Exchange I (ABIDE I)\citep{di_martino2014abide}, Attention Deficit Hyperactivity Disorder-200 (ADHD-200)\citep{milham2012adhd200}, and REST-meta-MDD~\citep{chen2022direct}.

The ABIDE I is a publicly available, multi-site dataset comprising resting-state functional magnetic resonance imaging (rs-fMRI) data~\citep{di_martino2014abide}. It includes individuals diagnosed with Autism Spectrum Disorder (ASD) and normal controls (NC). Preprocessing was performed using the Configurable Pipeline for the Analysis of Connectomes (CPAC)~\citep{craddock2013cpac}. Key preprocessing steps included band-pass filtering, global signal regression, and spatial normalization to a standard atlas.

The ADHD-200 dataset includes rs-fMRI scans from children and adolescents diagnosed with Attention-Deficit/Hyperactivity Disorder (ADHD) as well as typically developing controls (TDC). Preprocessing followed the Athena pipeline~\citep{bellec2017athena}, which included normalization to MNI space, nuisance regression, temporal filtering, and spatial smoothing. We further incorporated labeled test data from the ADHD-200 Global Competition to expand the dataset.

The REST-meta-MDD dataset, released by the Depression Imaging REsearch (DIRECT) consortium, aggregates rs-fMRI data from 25 research groups across 17 hospitals in China~\citep{chen2022direct}. The dataset contains both patients diagnosed with Major Depressive Disorder (MDD) and demographically matched healthy controls (NC). All imaging data were preprocessed using the standardized Data Processing Assistant for Resting-State fMRI (DPARSF) pipeline. Following previously established criteria, we excluded subjects with incomplete phenotypic records, poor spatial normalization, excessive head motion, or from sites with fewer than 10 participants per group~\citep{yan2019reduced}. 

Table~\ref{tab:datasets} summarizes the demographic characteristics of the three datasets. ADHD-200 primarily includes children and adolescents, whereas ABIDE I and REST-meta-MDD cover a broader age range extending into adulthood. ABIDE I and ADHD-200 show a clear male predominance, while REST-meta-MDD has a more balanced gender distribution. All datasets are collected from multiple sites, contributing to demographic and acquisition variability. ABIDE I includes 850 participants from 20 sites, ADHD-200 consists of 938 participants from 7 sites, and the REST-meta-MDD comprises 1,567 participants across 16 sites. While both ABIDE I and ADHD-200 include Full-Scale, Performance, and Verbal IQ scores, a proportion of these data are missing, with 22\% missing in ABIDE I and 26.3\% in ADHD-200. REST-meta-MDD includes years of education, which may serve as a proxy for cognitive and socioeconomic background.

\begin{table}[h]
\caption{Participant Demographics and Dataset Statistics for ABIDE I, ADHD-200 and Rest-meta-MDD}
\centering
\scriptsize  
\setlength{\tabcolsep}{4pt} 
\begin{tabular}{lccc}
\toprule 
\textbf{Dataset} & \textbf{ABIDE} & \textbf{ADHD200} & \textbf{Rest-meta-MDD}\\ 
\midrule 
Participants & ASD: 396, NC: 454 & ADHD: 450, TDC: 488 & MDD: 814, NC: 753 \\ 
Gender & M: 711, F: 139 & M: 588, F: 350 & M: 605, F: 962 \\ 
Sites & 20 & 7 & 16 \\ 
Age (years) & 6.47 - 58.0 & 7.09 - 26.31 & 18.0 - 65.0 \\ 
Education(years) &-&-&3.0- 23.0\\
IQ & FIQ, PIQ, VIQ & FIQ, PIQ, VIQ &-\\
\bottomrule 
\end{tabular}
\label{tab:datasets}
\end{table}

{
\subsection{Graph Definition in Brain Graph}
To construct the brain graph used in our framework, we define the nodes and edges for each dataset as follows.
For the ABIDE and ADHD-200 datasets, we used the Harvard-Oxford (HO) atlas \citep{ho_atlas}. Although it defines 111 ROIs, we excluded one unlabeled region, resulting in 110 ROIs. For the MDD datasets, we used the AAL atlas with its 116 ROIs\citep{tzourio2002automated}. Features were calculated as the Pearson correlation coefficients between every pair of ROIs within each atlas.

For the text encoder, we used BioBERT \citep{biobert}, a BERT-based language model pre-trained on biomedical literature. For each region, embeddings were generated by averaging the token embeddings from the last four hidden layers of the model that effectively capture nuanced contextual information. Edges between nodes were subsequently defined based on the similarity of these region embeddings. 

\subsection{Processing of Phenotypic Data}
To enable meaningful categorical comparisons, age was stratified into six groups to ensure balanced sample sizes. Phenotypic variables, including site, gender, and age group, were encoded using one-hot encoding. Given the critical role of intelligence quotient (IQ) in neurocognitive function \citep{nisbett2012intelligence}, particularly in neurodevelopmental disorders \citep{Wolff2022Autism}, we incorporated IQ data into the phenotypic features, accounting for variations in test types. IQ metrics considered included Full Scale IQ, Verbal IQ, and Performance IQ. The type of IQ test administered was also included to address potential variations in IQ distributions. For missing IQ data, we imputed the mean value for each test type to ensure consistency and maintain the accuracy of the dataset.
}

\subsection{Experimental Settings}

All experiments were conducted using 5 repetitions of 10-fold cross-validation to ensure robustness. Model performance was assessed using four commonly employed metrics: accuracy (ACC), area under the curve (AUC), specificity (SPE), and sensitivity (SEN). 

Models were implemented and evaluated within a PyTorch environment using an NVIDIA A800 GPU with 80 GB memory. Table~\ref{tab:config} summarizes the hyperparameter configurations for the two stages of the B2P-GL model across the ABIDE I, ADHD-200, and REST-meta-MDD datasets. The first stage adopts the ANR-GAT architecture, while the second stage is based on the HGCN framework. The table outlines dataset-specific settings for each stage, including the optimizer, learning rate, batch size, number of training epochs, hidden dimensions, and selected phenotypic variables.

\begin{table}[htbp]
\scriptsize
\centering
\caption{Model configuration and hyperparameters across different datasets}
\label{tab:config}
\begin{tabular}{lccc}
\toprule
\textbf{Parameters} & \textbf{ABIDE I} & \textbf{ADHD-200} & \textbf{Rest-meta-MDD} \\
\midrule
\multicolumn{4}{c}{\textbf{Stage 1}} \\
\midrule
optimizer & Adam & Adam & Adam \\
learning rate & $1.00\times10^{-4}$ & $1.00\times10^{-4}$ & $1.00\times10^{-4}$ \\
number of epoch & 200 & 200 & 200 \\
batch size & 32 & 32 & 32 \\
hidden channel & 512 & 256 & 512 \\
threshold& 0.6 & 0.5 & 0.6 \\
group number & 60 & 50 & 60 \\
\midrule
\multicolumn{4}{c}{\textbf{Stage 2}} \\
\midrule
optimizer & AdamW & AdamW & AdamW \\
learning rate & $1.00\times10^{-2}$ & $1.00\times10^{-2}$ & $1.00\times10^{-2}$ \\
weight decay & $5.00\times10^{-5}$ & $5.00\times10^{-5}$ & $5.00\times10^{-5}$ \\
number of epoch & 100 & 100 & 100 \\
hidden channel & 128 & 128 & 128 \\
top-k& 5 & 4 & 4 \\
condition selection & site, gender, age & site, gender & site, gender \\
phenotype selection & site, gender, age, IQ & site, IQ & site, education \\
\bottomrule
\end{tabular}
\end{table}

{
\subsection{Model Performance and Representation Analysis}
}
\begin{table*}[ht!]\footnotesize
\centering
\caption{Performance comparison of graph-based and non-graph models across multiple datasets. Models are divided into three categories: (1) non-graph neural networks, (2) graph classification using FC, and (3) node classification with phenotypic data. Proposed methods include ANR-GAT and B2P-GL. Best results per metric and dataset are bolded.}
\label{tab:performance}
\resizebox{\textwidth}{!}{%
\begin{tabular}{l|cccc|cccc|cccc}
\hline
\multirow{2}{*}{\textbf{Model}}
& \multicolumn{4}{c|}{\textbf{ABIDE I}}
& \multicolumn{4}{c|}{\textbf{ADHD-200}}
& \multicolumn{4}{c}{\textbf{REST-meta-MDD}} \\
\cline{2-13}
& \textbf{ACC} & \textbf{AUC} & \textbf{SPE} & \textbf{SEN}
& \textbf{ACC} & \textbf{AUC} & \textbf{SPE} & \textbf{SEN}
& \textbf{ACC} & \textbf{AUC} & \textbf{SPE} & \textbf{SEN} \\
&  (\%, $\uparrow$) &  (\%, $\uparrow$) &  (\%, $\uparrow$) &  (\%, $\uparrow$)
&  (\%, $\uparrow$) &  (\%, $\uparrow$) &  (\%, $\uparrow$) &  (\%, $\uparrow$)
&  (\%, $\uparrow$) &  (\%, $\uparrow$) &  (\%, $\uparrow$) &  (\%, $\uparrow$) \\
\hline
\multicolumn{13}{c}{\textbf{Non-Graph Neural Networks}} \\
\hline
ViT 
& 66.8 $\pm$ 1.4 & 66.2 $\pm$ 1.7 & 74.5 $\pm$ 3.9 & 58.0 $\pm$ 6.5
& 68.3 $\pm$ 0.7 & 62.8 $\pm$ 1.6 & 85.6 $\pm$ 2.2 & 40.0 $\pm$ 5.3
& 63.8 $\pm$ 0.5 & 63.6 $\pm$ 0.6 & 57.4 $\pm$ 4.8 & 69.7 $\pm$ 3.9 \\
CNN
& 69.0 $\pm$ 0.5 & 68.7 $\pm$ 0.4 & 72.0 $\pm$ 4.6 & 65.5 $\pm$ 4.7
& 66.0 $\pm$ 0.5 & 58.1 $\pm$ 1.1 & 91.1 $\pm$ 1.7 & 25.1 $\pm$ 3.7
& 65.1 $\pm$ 1.5 & 65.1 $\pm$ 1.4 & 63.8 $\pm$ 3.2 & 66.3 $\pm$ 5.3 \\
BrainNetCNN
& \textbf{72.2 $\pm$ 1.1} & \textbf{72.0 $\pm$ 1.3} & \textbf{74.9 $\pm$ 2.1} & \textbf{69.1 $\pm$ 4.2}
& 66.3 $\pm$ 0.6 & 59.6 $\pm$ 1.5 & \textbf{87.8 $\pm$ 2.3} & 31.3 $\pm$ 5.3
& \textbf{66.6 $\pm$ 1.0} & \textbf{66.6 $\pm$ 1.0} & \textbf{66.1 $\pm$ 2.2} & \textbf{67.1 $\pm$ 2.7} \\
BrainNetTransformer
& 69.5 $\pm$ 0.1 & 69.1 $\pm$ 0.3 & 73.7 $\pm$ 1.4 & 64.5 $\pm$ 2.1
& \textbf{68.7 $\pm$ 0.3} & \textbf{64.1 $\pm$ 1.0} & 83.2 $\pm$ 1.8 & \textbf{45.0 $\pm$ 3.7}
& 63.9 $\pm$ 0.1 & 63.8 $\pm$ 0.2 & 60.6 $\pm$ 4.0 & 67.0 $\pm$ 3.7 \\
\hline
\multicolumn{13}{c}{\textbf{Graph Classification Models}} \\
\hline
GAT
& 69.7 $\pm$ 0.3 & 69.1 $\pm$ 0.2 & 77.5 $\pm$ 1.1 & 60.8 $\pm$ 0.6
& 60.1 $\pm$ 0.4 & 59.8 $\pm$ 0.3 & \textbf{66.0 $\pm$ 1.7} & 53.6 $\pm$ 0.3
& 65.1 $\pm$ 0.2 & 64.9 $\pm$ 0.2 & 60.3 $\pm$ 0.4 & 69.4 $\pm$ 0.5 \\
BrainGNN
& 67.0 $\pm$ 0.4 & 65.3 $\pm$ 0.7 & 74.3 $\pm$ 2.4 & 56.5 $\pm$ 2.2
& 61.6 $\pm$ 0.7 & 60.6 $\pm$ 0.8 & 61.5 $\pm$ 1.8 & 61.6 $\pm$ 1.1
& 64.0 $\pm$ 0.1 & 65.0 $\pm$ 0.2 & 64.3 $\pm$ 0.9 & 63.8 $\pm$ 0.8 \\
TSEN
& 70.1 $\pm$ 0.3 & 69.5 $\pm$ 0.4 & 79.3 $\pm$ 2.0 & 59.8 $\pm$ 2.4
& 62.2 $\pm$ 0.6 & 62.4 $\pm$ 0.6 & 57.4 $\pm$ 3.8 & \textbf{67.4 $\pm$ 4.1}
& \textbf{66.7 $\pm$ 0.7} & 66.6 $\pm$ 0.7 & 66.6 $\pm$ 3.2 & \textbf{66.7 $\pm$ 2.8} \\
\textbf{ANR-GAT (ours)}
& \textbf{73.7 $\pm$ 0.6} & \textbf{73.2 $\pm$ 0.6} & \textbf{80.5 $\pm$ 1.8} & \textbf{65.9 $\pm$ 2.1}
& \textbf{65.6 $\pm$ 0.6} & \textbf{65.5 $\pm$ 0.6} & 64.9 $\pm$ 1.3 & 63.4 $\pm$ 3.8
& 66.6 $\pm$ 0.2 & \textbf{66.8 $\pm$ 0.2} & \textbf{71.3 $\pm$ 2.0} & 62.3 $\pm$ 2.1 \\
\hline
\multicolumn{13}{c}{\textbf{Node Classification Models}} \\
\hline
Population-GCN
& 71.1 $\pm$ 0.2 & 73.9 $\pm$ 0.5 & 81.3 $\pm$ 0.6 & 59.3 $\pm$ 0.9
& 62.9 $\pm$ 1.1 & 62.6 $\pm$ 2.0 & 61.6 $\pm$ 3.2 & 64.1 $\pm$ 1.8
& 67.2 $\pm$ 0.6 & 69.4 $\pm$ 0.1 & 62.3 $\pm$ 2.9 & 71.8 $\pm$ 1.5 \\
EV-GCN
& 71.6 $\pm$ 0.3 & 72.7 $\pm$ 0.4 & 78.6 $\pm$ 1.7 & 63.6 $\pm$ 1.6
& \textbf{67.4 $\pm$ 0.1} & 66.9 $\pm$ 0.2 & 68.8 $\pm$ 2.9 & 66.1 $\pm$ 2.4
& 66.5 $\pm$ 0.3 & \textbf{67.4 $\pm$ 0.4} & 60.0 $\pm$ 1.9 & 72.5 $\pm$ 1.5 \\
LG-GNN
& 80.1 $\pm$ 0.8 & 81.3 $\pm$ 1.2 & 81.6 $\pm$ 3.0 & \textbf{78.8 $\pm$ 2.7}
& 65.9 $\pm$ 0.9 & 65.3 $\pm$ 1.6 & 68.2 $\pm$ 3.0 & 63.3 $\pm$ 4.1
& 79.9 $\pm$ 0.4 & \textbf{90.6 $\pm$ 0.2} & \textbf{91.2 $\pm$ 0.4} & 69.2 $\pm$ 1.1 \\
\textbf{B2P-GL (ours)}
& \textbf{83.4 $\pm$ 0.5} & \textbf{83.0 $\pm$ 0.3} & \textbf{88.9 $\pm$ 0.5} & 77.1 $\pm$ 0.1
& \textbf{78.6 $\pm$ 0.4} & \textbf{78.8 $\pm$ 0.4} & \textbf{82.3 $\pm$ 0.8} & \textbf{75.2 $\pm$ 1.5}
& \textbf{82.0 $\pm$ 0.3} & 82.0 $\pm$ 0.3 & 81.3 $\pm$ 1.5 & \textbf{82.6 $\pm$ 1.1} \\
\hline
\end{tabular}%
}
\end{table*}

To comprehensively assess the performance of our proposed models, we conducted a systematic comparison across three major types of architectures: (1) non-graph neural networks, which treat the FC matrix as an image input; (2) graph classification models, which operate on individual-level FC graphs; and (3) node classification models, which leverage population graphs constructed from individual embeddings and phenotypic metadata.

For non-graph baselines, we considered four representative architectures: ViT\citep{dosovitskiy2020image}, CNN\citep{krizhevsky2017imagenet}, BrainNetCNN\citep{kawahara2017brainnetcnn}, and BrainNetTransformer\citep{kan2022brain}, which directly process FC matrices as images. These models provide a strong baseline for evaluating whether graph structure offers added benefit over traditional 2D processing pipelines.

In the graph-based category, we evaluated both brain and population level approaches. For brain graph classification, we included GAT \citep{velivckovic2017graph}, BrainGNN \citep{braingnn}, and TSEN \citep{tsen}, which operate on individual functional connectivity graphs and aim to predict individual-specific diagnostic labels. In comparison, population graph models such as Population-GCN \citep{populationgcn}, EV-GCN \citep{evgcn}, and LG-GNN \citep{lggnn} integrate non-imaging variables and are trained in a transductive setting.

Our proposed methods include ANR-GAT, which enhances brain network representation learning through attention graph network and node reassignment strategies, and B2P-GL, which integrates individual brain features with condition-based population modeling and multimodal phenotypic fusion.

As shown in Table~\ref{tab:performance}, our proposed models outperform existing baselines across most datasets and evaluation metrics. Among non-graph models, BrainNetCNN and BrainNetTransformer exhibit solid results, particularly on ADHD-200 dataset. However, their inability to explicitly model relational structure between brain regions or individuals limits their generalizability.

Among individual-level graph models, ANR-GAT achieves the best accuracy on ABIDE I and ADHD-200. On ABIDE I, it improves accuracy and AUC by 3.6\% and 3.7\% over the previous best (TSEN), and also yields gains in both specificity and sensitivity. On ADHD-200, it raises AUC by 3.1\% and sensitivity by 4.0\%, showing stronger detection of neurodevelopmental patterns. Although its performance on REST-meta-MDD is slightly below TSEN, it is more stable with lower variance.

At the population level, B2P-GL achieves top or second-best results on all datasets. On ABIDE I, it improves accuracy by 3.3\% and specificity by 7.3\% over LG-GNN. On ADHD-200, the model improves accuracy by 12.7\% and sensitivity by 6.9\% over LG-GNN. On REST-meta-MDD, it improves accuracy by 2.1\% and achieves the highest sensitivity, exceeding LG-GNN by 13.4\%.

\begin{figure}[h]
    \centering
    \includegraphics[width=1\linewidth]{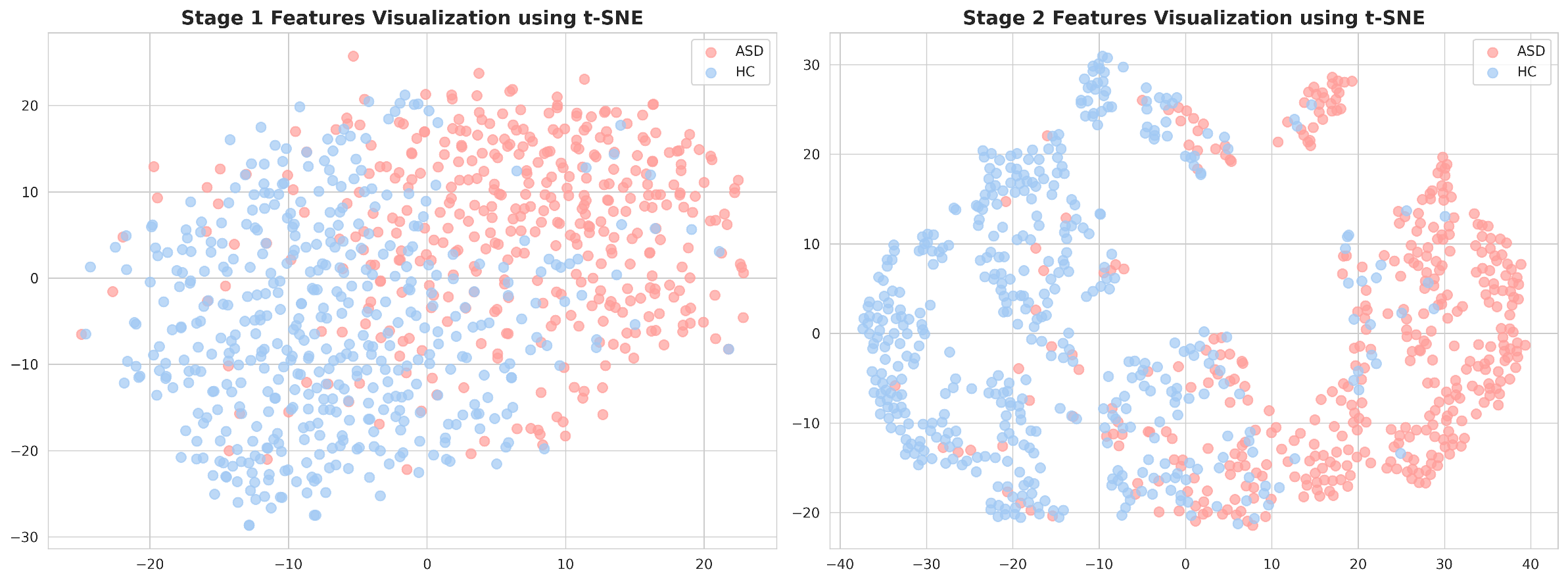}
    \caption{Visualization of brain feature embeddings using t-SNE. Compared with stage 1 embeddings derived from individual graphs and stage 2 embeddings updated through HGCN without phenotypic fusion.}
    \label{fig:feature_visualization}
\end{figure}

To provide a more direct demonstration of the effectiveness of condition-based graph construction and HGCN in stage 2, we further visualized the feature embeddings using t-SNE. We compared the brain feature embeddings learned from stage 1 with those updated through HGCN without phenotypic fusion. In Figure \ref{fig:feature_visualization}, the class distance increased markedly from 0.6769 in stage 1 to 16.2326 in stage 2, indicating that the embeddings obtained in the population graph capture more disease-relevant information beyond site and demographic effects.

These results highlight the effectiveness of our two-stage strategy, which first refines individual brain representations and then leverages condition-based population modeling. By jointly incorporating functional connectivity and phenotypic features, our models capture individual variability while remaining robust across disorders with distinct neurobiological signatures.

\subsection{Analysis of Brain Representative Learning}

The semantic similarity used for edge construction and the adaptive node reassignment mechanism are two key components in the brain representation learning stage. In this section, we perform targeted evaluations of both modules to assess their individual contributions and effectiveness in improving the performance of brain graph representations.

\subsubsection{Effectiveness of Semantic Similarity in Edge}

To evaluate the effectiveness of brain region similarity as an edge definition compared to traditional approaches in brain networks, we conducted a comprehensive analysis using the following edge definitions for the brain graph representation learning: (1) Pearson Correlation with global regression (GR), (2) Pearson Correlation without global regression (NGR), (3) Partial Correlation with GR, (4) Transfer Entropy with GR, and (5) Semantic Similarity of brain regions. Here, global regression refers to the removal of the global mean time series from all brain region signals before computing Pearson or partial correlations.

The results in Table~\ref{tab:ablation_embeddings} indicate that using semantic similarity for edge definition leads to the best overall performance. Compared to the widely used Pearson correlation with global regression, it yields a 0.7\% increase in both accuracy and AUC. This performance gain suggests that semantic similarity enriches the informational content of the brain graph representation by embedding prior anatomical and functional knowledge into the graph structure. Unlike conventional approaches that rely on signal-based correlations, the semantic approach introduces structurally meaningful connections that are less sensitive to noise and inter-individual variability. Overall, the results highlight the value of incorporating language-informed brain priors to complement and enhance traditional neuroimaging-derived connectivity measures.

\begin{table}[h]
\scriptsize
\centering
\caption{Performance comparison of different edge definitions used in brain graph construction.}
\setlength{\tabcolsep}{1.3mm}{
\begin{tabular}{lcccc}
\toprule
\textbf{Edge Type} & \textbf{ACC (\%, $\uparrow$)} & \textbf{AUC (\%, $\uparrow$)} & \textbf{SEN (\%, $\uparrow$)} & \textbf{SPE (\%, $\uparrow$)} \\ 
\midrule
Pearson Correlation (GR) & 73.0 $\pm$ 0.7 & 72.5 $\pm$ 0.7 & 80.1 $\pm$ 2.6 & 64.9 $\pm$ 2.5 \\
Pearson Correlation (NGR) & 72.6 $\pm$ 0.5 & 71.6 $\pm$ 0.5 & \textbf{86.2 $\pm$ 1.7} & 56.9 $\pm$ 1.6 \\
Partial Correlation (GR) & 72.4 $\pm$ 0.6 & 72.1 $\pm$ 0.7 & 77.0 $\pm$ 0.9 & \textbf{67.2 $\pm$ 1.9} \\
Transfer Entropy (GR) & 73.0 $\pm$ 1.0 & 72.5 $\pm$ 1.1 & 79.5 $\pm$ 1.7 & 65.5 $\pm$ 3.6 \\
\textbf{Semantic Similarity} & \textbf{73.7 $\pm$ 0.6} & \textbf{73.2 $\pm$ 0.6} & 80.5 $\pm$ 1.8 & 65.9 $\pm$ 2.1 \\
\bottomrule
\end{tabular}
}
\label{tab:ablation_embeddings}
\end{table}

In addition, we further evaluated two strategies for computing semantic similarity: (1) lexical embeddings of region names and (2) contextualized embeddings incorporating their concepts. Subsequently, we compared BERT-base \citep{DBLP:journals/corr/abs-1810-04805}, a pre-trained language model designed for general tasks, and BioBERT, a biomedical domain-adapted variant. 

As shown in Table~\ref{tab:semantic_similarity}, incorporating contextualized concepts led to consistent performance gains. For BERT-base, concept-based embeddings improved accuracy by 0.6\%. BioBERT also benefited from concept augmentation, with a 0.4\% increase. When comparing models under the same input setting (i.e., concepts), BioBERT outperformed BERT-base by 0.3\% in accuracy and 0.4\% in AUC, indicating its advantage as a more suitable choice for neurological applications. These results highlight the importance of both domain-adapted language models and concept-level augmentation for capturing meaningful semantic similarity between brain regions.

\begin{table}[h]
\scriptsize
\centering
\caption{Comparison of brain region similarity strategies using BERT and BioBERT embeddings. “Names” uses lexical embeddings of region names, while “Concepts” incorporates additional anatomical and functional context.}
\vskip 0.5em
\label{tab:semantic_similarity}
\renewcommand{\arraystretch}{1.15}
\setlength{\tabcolsep}{2.1mm}{
\begin{tabular}{@{}llcccc@{}}
\toprule
\textbf{Scope} & \textbf{Model} & \textbf{ACC (\%, $\uparrow$)} & \textbf{AUC (\%, $\uparrow$)} & \textbf{SPE (\%, $\uparrow$)} & \textbf{SEN (\%, $\uparrow$)} \\
\midrule
\multirow{2}{*}{Names} & BERT     & 72.8 $\pm$ 0.6 & 72.3 $\pm$ 0.6 & 80.2 $\pm$ 1.8 & 64.4 $\pm$ 2.4 \\
                       & BioBERT  & 73.3 $\pm$ 0.8 & 72.9 $\pm$ 0.8 & 79.2 $\pm$ 1.5 & \textbf{66.5 $\pm$ 1.6} \\
\addlinespace
\multirow{2}{*}{Concepts} & BERT     & 73.4 $\pm$ 0.5 & 72.8 $\pm$ 0.6 & \textbf{81.7 $\pm$ 2.0} & 64.0 $\pm$ 2.5 \\
                                & BioBERT  & \textbf{73.7 $\pm$ 0.6} & \textbf{73.2 $\pm$ 0.6} & 80.5 $\pm$ 1.8 & 65.9 $\pm$ 2.1 \\
\bottomrule
\end{tabular}
}
\end{table}

\subsubsection{Effectiveness of Node Reassignment and Group Size Sensitivity}
When training the brain graphs, our adaptive node reassignment module dynamically refines brain region assignments based on input data, enabling a more flexible and personalized representation of functional brain networks. To assess its effectiveness, we performed an ablation study comparing three configurations: (i) ANR-GAT without the node reassignment module, (ii) a variant where the reassignment mechanism is replaced with a WGAT layer, and (iii) the full proposed ANR-GAT model.

As shown in Table~\ref{tab:node_reassignment_ablation}, removing the node reassignment module leads to a noticeable drop in model performance, with accuracy and AUC decreasing to 70.6\% and 70.4\%, respectively. Replacing the adaptive reassignment mechanism with a WGAT layer yields some improvement over the removal case, indicating that incorporating structural grouping is beneficial. However, this variant still falls short of the full ANR-GAT model, which achieves the highest accuracy of 73.7\% and AUC of 73.2\%. These results demonstrate that our adaptive node reassignment strategy plays a critical role in learning flexible and individualized brain region groupings while dynamically updating semantic edges based on empirical functional connectivity.

\begin{table}[ht]
\scriptsize
\centering
\caption{Ablation study of the node reassignment module in ANR-GAT. Top: complete removal. Middle: replacement with a WGAT layer. Bottom: our full proposed model.}
\setlength{\tabcolsep}{1.6mm}{
\begin{tabular}{lcccc}
\toprule
\textbf{Model Variant} & \textbf{ACC (\%, $\uparrow$)} & \textbf{AUC (\%, $\uparrow$)} & \textbf{SEP (\%, $\uparrow$)} & \textbf{SEN (\%, $\uparrow$)} \\
\midrule
w/o Node Reassignment & 70.6 $\pm$ 0.5 & 70.4 $\pm$ 0.4 & 72.2 $\pm$ 4.0 & 68.7 $\pm$ 3.9 \\
WGAT Replacement & 72.1 $\pm$ 0.1 & 71.2 $\pm$ 0.1 & 84.0 $\pm$ 2.2 & 58.4 $\pm$ 2.3 \\
Full ANR-GAT & \textbf{73.7 $\pm$ 0.6} & \textbf{73.2 $\pm$ 0.6} & \textbf{80.6 $\pm$ 1.8} & \textbf{65.9 $\pm$ 2.1} \\
\bottomrule
\end{tabular}
}
\label{tab:node_reassignment_ablation}
\end{table}

To further examine the sensitivity of our model to the projection space configuration, we evaluated the impact of varying the predefined group size in the feature-to-group mapping component on the ABIDE dataset. The group size determines the granularity of the reassignment space, influencing the capacity of the model to represent diverse node features. Larger group sizes can capture fine-grained clusters but may introduce redundancy, whereas smaller sizes might oversimplify the structural patterns. Our method addresses this trade-off by leveraging a hard assignment mechanism that treats the group size as an upper bound rather than a strict constraint. Groups are dynamically selected based on the data distribution, and only a subset of them is utilized depending on relevance. Empty groups naturally emerge from this process as part of adaptive sparsity, whereby the model prunes uninformative groups. This not only improves computational efficiency but also mitigates overfit. To quantify this behavior, we define the \textit{group assignment ratio} as the proportion of non-empty groups.

\begin{figure}[h]
    \centering
    \includegraphics[width=1\linewidth]{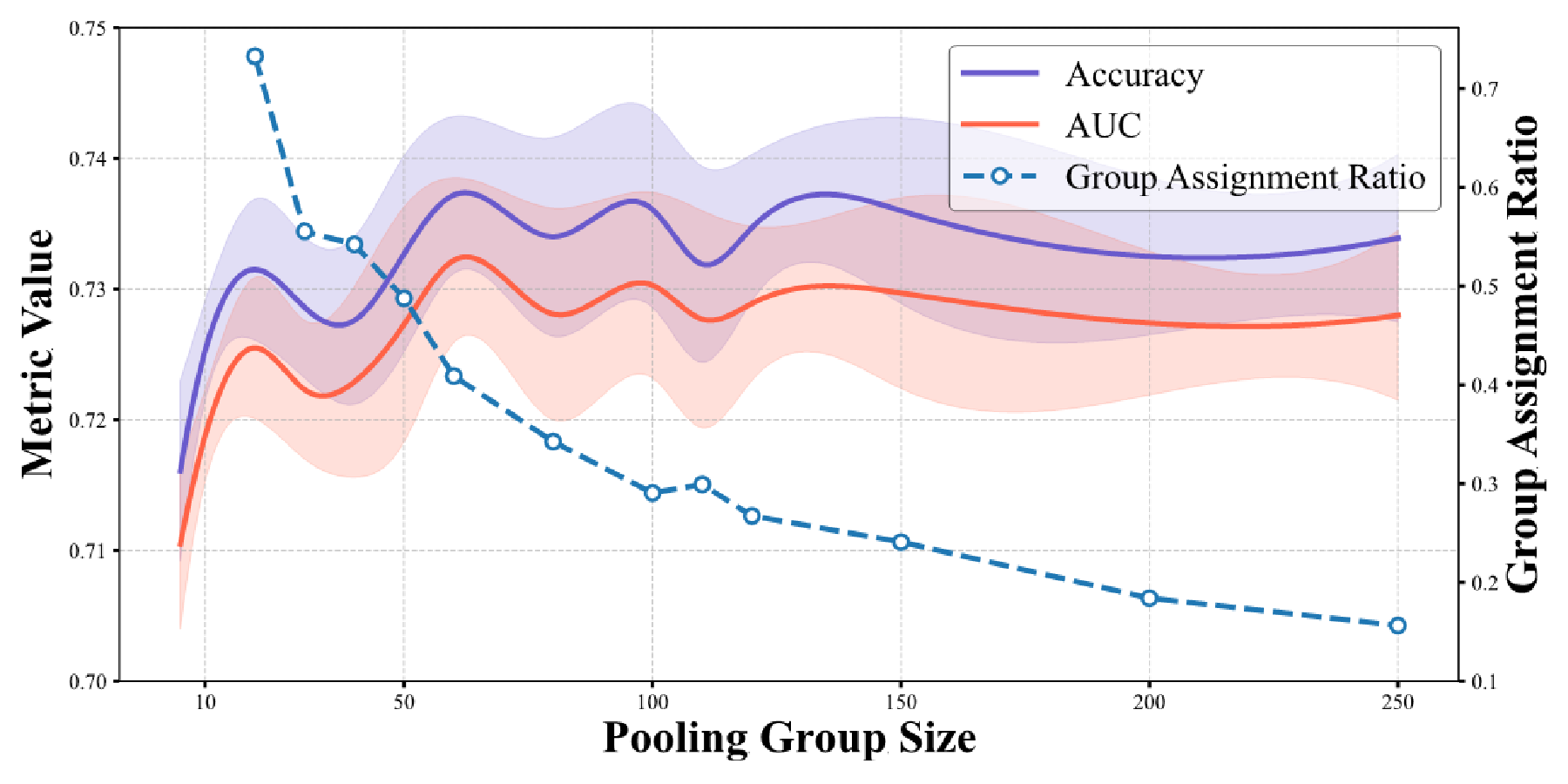}
    \caption{Performance trends in ACC and AUC alongside the group assignment ratio across varying group sizes. The left y-axis shows metric values for ACC and AUC, while the right y-axis represents the group assignment ratio. The results demonstrate the influence of group size on model performance and the actual group assignment numbers using hard assignment.}
    \label{fig:group performance}
\end{figure}

As illustrated in Figure~\ref{fig:group performance}, the model achieves optimal performance when the group size lies between 60 and 100, given the original node size of 110. Within this range, the model effectively balances expressivity and generalization, capturing discriminative group-level patterns while preserving individual region characteristics. Beyond this interval, although the predefined group size may exceed the original node count, the number of actively used groups grows only marginally. This observation suggests a saturation regime, in which expanding the group size no longer enhances the model’s representational capacity or contributes meaningfully to the encoding of structural information.

{
\subsubsection{Parameter Sensitivity and Computational Trade-off}
The computational cost in brain representation learning is primarily influenced by two factors: the hidden dimension of the GAT layers and the semantic similarity threshold used to construct the brain graph. These parameters jointly determine the balance between model expressiveness, graph sparsity, and computational efficiency. In what follows, we provide a detailed analysis of the threshold, which directly affects both graph structure and computational cost.

The semantic similarity threshold controls the density of the constructed brain graph. A smaller value preserves more connections, leading to higher GPU memory usage and longer training time, and may introduce noisy or redundant connections that weaken the quality of learned representations. Conversely, a larger threshold prunes more edges, reducing the computational cost but potentially discarding informative relationships among brain regions. To examine this trade-off, we conducted experiments with threshold values from 0.4 to 0.8 in the brain representation learning stage across different datasets, each using its best-performing configuration. As shown in Figure~\ref{fig:threshold}, moderate thresholds consistently yield the best results, capturing meaningful connectivity while maintaining computational efficiency.

\begin{figure}[h]
    \centering
    \includegraphics[width=1\linewidth]{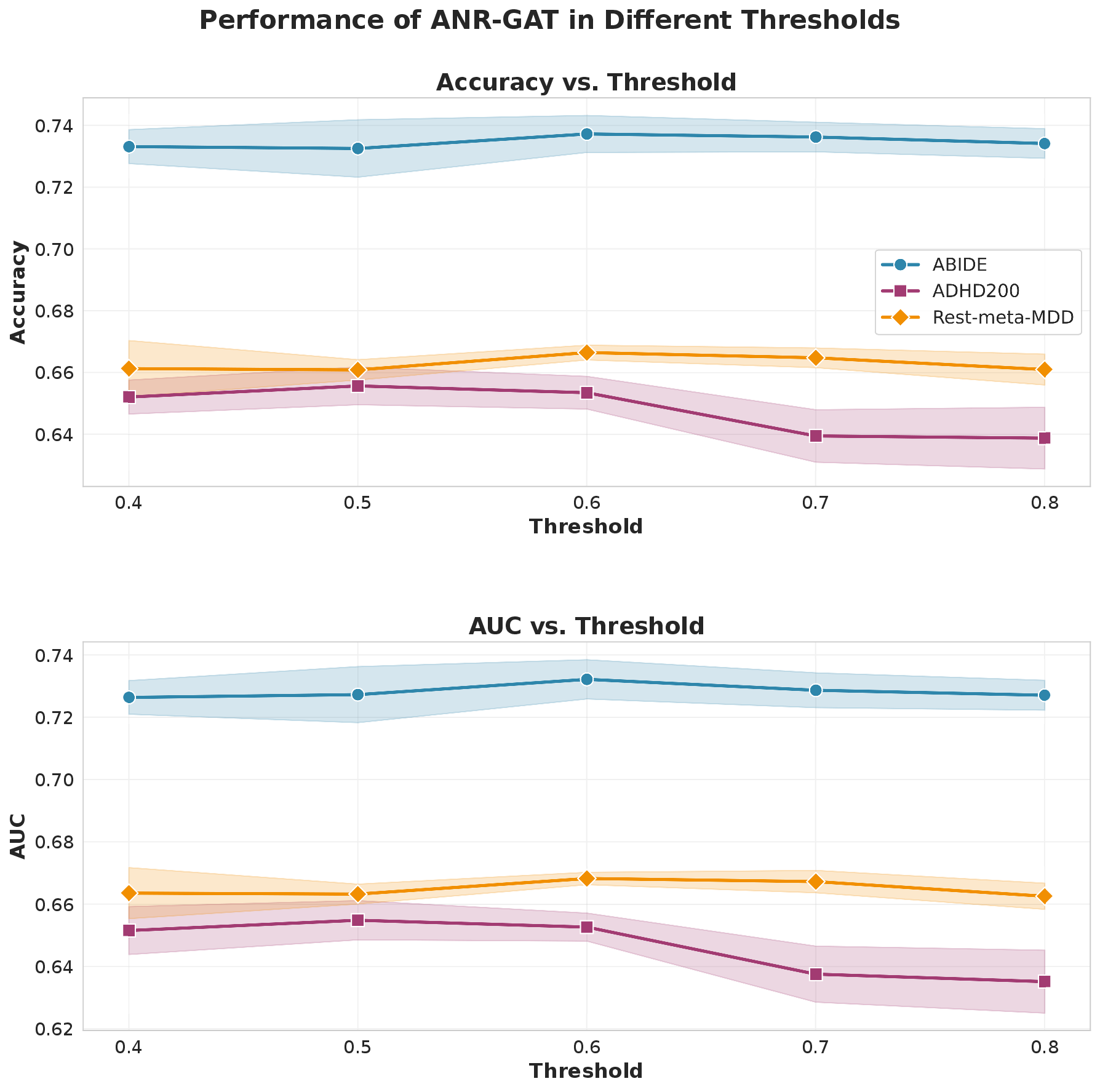}
    \caption{Accuracy and AUC performance of ANR-GAT model with varying threshold values (0.4-0.8) on ABIDE, ADHD200, and Rest-meta-MDD datasets.}
    \label{fig:threshold}
\end{figure}

While the other experiment parameters are also primarily selected by model performance, the corresponding GPU utilization and training time under the adopted settings are reported to assess computational feasibility. As shown in Table~\ref{tab:stage1_params}, GPU memory usage is well within the available resources, and the per-epoch training time remains consistently reasonable across datasets.

\begin{table}[h]
\renewcommand{\arraystretch}{1.15}
\scriptsize
\centering
\caption{Stage 1 parameter settings and measured computational cost (maximum GPU utilization recorded in wandb, and training time per epoch).}
\label{tab:stage1_params}
\setlength{\tabcolsep}{1.7mm}{
\begin{tabular}{lcccc}
\toprule
Dataset & Hidden channel & Threshold & 
\begin{tabular}{@{}c@{}}Max GPU \\ utilization (\%)\end{tabular} & 
\begin{tabular}{@{}c@{}}Training time \\ / epoch (s)\end{tabular} \\
\midrule
ABIDE & 512 & 0.60 & 59.53 & 1.18--1.24 \\
ADHD-200 & 256 & 0.50 & 47.98 & 0.85--0.89 \\
Rest-meta-MDD & 512 & 0.60 & 63.34 & 1.94--2.00 \\
\bottomrule
\end{tabular}
}
\end{table}

These observations confirm that the adopted configurations achieve a reasonable balance between model performance and computational efficiency.

\subsection{Analysis of Population Disease Diagnosis}
The condition-based population graph construction and the integration of phenotypic information represent two essential components of the population-level disease diagnosis stage. In this section, we evaluate the contribution of phenotypic variables to both the construction of the population graph and the enhancement of individual node features through feature fusion.

\subsubsection{Impact of Cross-Condition Population Graph}
We evaluated the impact of site, age, and gender conditions on edge generation in the population network. Edge selection dynamically adjusted the top-$k$ edges for each condition, ensuring a proportional total edge count. 
As shown in Table \ref{tab:condition}, individual conditions produced comparable results, with gender achieving slightly higher accuracy. Pairwise combinations improved performance incrementally, while integrating all three conditions yielded the best results, with a 1.6\% increase in accuracy and a 1.5\% increase in AUC. These results demonstrate the importance of leveraging multiple conditions for robust edge generation in brain disorder analysis.

\begin{table}[h]
\caption{Performance comparison using different combinations of site, age, and gender conditions for edge construction.}
\scriptsize
\centering
\setlength{\tabcolsep}{1.7mm}{
\begin{tabular}{cccccccc}
\toprule
 \textbf{Site} & \textbf{Age} & \textbf{Gender} & \textbf{ACC (\%, $\uparrow$)} & \textbf{AUC (\%, $\uparrow$)} & \textbf{SPE (\%, $\uparrow$)} & \textbf{SEN (\%, $\uparrow$)} \\ 
\midrule
 \checkmark &  &  & 82.1 $\pm$ 0.4 & 81.8 $\pm$ 0.5 & 86.9 $\pm$ 0.7 & 76.7 $\pm$ 1.6 \\
  & \checkmark &  & 82.1 $\pm$ 0.3 & 81.8 $\pm$ 0.4 & 86.3 $\pm$ 0.1 & 77.3 $\pm$ 0.8 \\
  &  & \checkmark & 82.5 $\pm$ 0.1 & 82.1 $\pm$ 0.2 & 87.7 $\pm$ 0.6 & 76.6 $\pm$ 1.0 \\
 \checkmark & \checkmark &  & 82.8 $\pm$ 0.9 & 82.4 $\pm$ 1.0 & 87.6 $\pm$ 0.3 & 77.3 $\pm$ 1.8 \\
 \checkmark &  & \checkmark & 82.7 $\pm$ 0.5 & 82.5 $\pm$ 0.6 & 86.0 $\pm$ 0.9 & \textbf{79.0 $\pm$ 1.8} \\
  & \checkmark & \checkmark & 82.6 $\pm$ 0.5 & 82.3 $\pm$ 0.5 & 86.2 $\pm$ 1.3 & 78.4 $\pm$ 0.4 \\
 \checkmark & \checkmark & \checkmark & \textbf{83.4 $\pm$ 0.5} & \textbf{83.0 $\pm$ 0.3} & \textbf{88.9 $\pm$ 0.5} & 77.1 $\pm$ 0.1 \\
\bottomrule
\end{tabular}
}
\label{tab:condition}
\end{table}

To investigate the impact of condition-based graph construction and the corresponding heterogeneous modeling strategy, we compared three edge generation and graph learning methods.

Firstly, we constructed a phenotypic-brain similarity graph, in which edges are based on pure similarity of phenotypic and brain features, without differentiating between specific demographic categories. Since no condition type is encoded in the graph structure, we apply a standard GCN that treats all edges uniformly. This serves as our baseline, where neither edge generation nor model training is aware of demographic distinctions. 

We then evaluated the condition-based population graph, where edges are constructed based on shared categorical attributes. 
To assess the benefit of allowing information exchange across demographic boundaries, we compared two variants: same-category edges, which connect the same category in one condition, and cross-category edges, which also allow links across different categories. The HGCN was employed to effectively model the multiple condition-specific edge types by assigning distinct transformation functions to each relation.

As shown in Table~\ref{tab:edge_model_comparison}, the GCN trained on the phenotypic-brain similarity graph achieves moderate performance. In contrast, our HGCN applied to the same condition-based graph yields clear performance gains. Notably, when cross-category edges are included, allowing information to flow across demographic boundaries, our model achieves the best results, with a 1.4\% improvement in accuracy and 1.3\% in AUC over the baseline.  These experiments highlight the advantage of leveraging inter-demographic relations for more effective population-level graph learning.

\begin{table}[ht]
\centering
\scriptsize
\caption{Performance comparison of different edge construction strategies and model architectures. The Phenotypic-Brain Similarity graph uses a standard GCN without condition-based modeling.}
\label{tab:edge_model_comparison}
\setlength{\tabcolsep}{1.5mm}{
\begin{tabular}{lcccc}
\toprule
\textbf{Edge Type / Model} & \textbf{ACC(\%, $\uparrow$)} & \textbf{AUC(\%, $\uparrow$)} & \textbf{SPE (\%, $\uparrow$)} & \textbf{SEN (\%, $\uparrow$)} \\ 
\midrule
Phenotypic-Brain & 82.0 $\pm$ 0.5 & 81.7 $\pm$ 0.5 & 85.3 $\pm$ 1.9 & \textbf{78.1 $\pm$ 1.9} \\
Similarity (GCN) & & & & \\[0.5ex]

Condition-based: & 82.0 $\pm$ 0.5 & 81.7 $\pm$ 0.6 & 86.6 $\pm$ 0.7 & 76.8 $\pm$ 1.5 \\
Same-category (HGCN) & & & & \\[0.5ex]
Condition-based: & \textbf{83.4 $\pm$ 0.4} & \textbf{83.0 $\pm$ 0.3} & \textbf{88.9 $\pm$ 0.5} & 77.1 $\pm$ 0.1 \\
Different-category (HGCN) & & & & \\
\bottomrule
\end{tabular}
}
\end{table}

\subsubsection{Impact of Phenotypic Feature Fusion}

To assess the contribution of phenotypic information to ASD classification, we investigated how fusing individual-level attributes with brain representations influences model performance on the ABIDE I dataset. The phenotypic features considered in this analysis include site, age, gender, and IQ. 

As shown in Table~\ref{table:phenotypic}, integrating individual features leads to varying degrees of performance improvement. Among all features, IQ yields the most notable gain, substantially enhancing classification accuracy and AUC. Combining multiple features results in further improvements: both pairwise and triplet combinations outperform single-feature fusion, indicating that these attributes provide complementary information. The best performance is achieved when all four features are fused, leading to a 6.3\% increase in accuracy and a 6.1\% gain in AUC compared to using IQ alone. These findings highlight the value of incorporating diverse phenotypic sources. By capturing individual differences beyond imaging data, the model benefits from a more holistic individual representation, ultimately improving its ability to detect disorder-related patterns.

\begin{table}[h]
\caption{Impact of different phenotypic feature combinations on performance when fused with brain representations. The first row shows results without phenotypic fusion.}
\scriptsize
\centering
\setlength{\tabcolsep}{1.3mm}{
\begin{tabular}{cccccccc}
\toprule
 \textbf{Site} & \textbf{Age} & \textbf{Gender} & \textbf{IQ} & \textbf{ACC(\%, $\uparrow$)} & \textbf{AUC(\%, $\uparrow$)} & \textbf{SPE (\%, $\uparrow$)} & \textbf{SEN (\%, $\uparrow$)} \\ 
\midrule
 &  &  &  & 77.1 $\pm$ 0.4  & 76.7 $\pm$ 0.4 & 82.1 $\pm$ 1.4 & 71.3 $\pm$ 1.1 \\
\checkmark &  &  &  & 77.3 $\pm$ 0.4 & 77.1 $\pm$ 0.5 & 80.1 $\pm$ 0.7 & 74.2 $\pm$ 1.2 \\
\checkmark & \checkmark &  &  & 77.6 $\pm$ 0.9 & 77.3 $\pm$ 0.9 & 82.3 $\pm$ 0.9 & 72.2 $\pm$ 1.1 \\
\checkmark &  & \checkmark &  & 77.3 $\pm$ 0.7 & 77.0 $\pm$ 0.8 & 81.2 $\pm$ 0.8 & 72.8 $\pm$ 2.0 \\
\checkmark &  &  & \checkmark & 82.1 $\pm$ 0.6 & 81.9 $\pm$ 0.6 & 84.8 $\pm$ 0.9 & \textbf{79.1 $\pm$ 1.0} \\
\checkmark & \checkmark & \checkmark &  & 77.8 $\pm$ 0.5 & 77.9 $\pm$ 0.6 & 81.5 $\pm$ 2.0 & 74.4 $\pm$ 1.2 \\
\checkmark & \checkmark & \checkmark & \checkmark & \textbf{83.4 $\pm$ 0.5} & \textbf{83.0 $\pm$ 0.3} & \textbf{88.9 $\pm$ 0.5} & 77.1 $\pm$ 0.1 \\

\bottomrule
\end{tabular}
}
\label{table:phenotypic}
\end{table}

To evaluate the effectiveness of integrating phenotypic features, we conduct a series of fusion experiments using various strategies. Specifically, we compare four approaches: (i) add fusion, which combines features through element-wise addition; (ii) concat fusion, which concatenates feature vectors before transformation; (iii) attention fusion, which assigns adaptive weights to different modalities using a learned attention mechanism; and (iv) gated fusion (ours), which employs a gating module to selectively control the contribution of each feature dimension.

As shown in Table~\ref{tab:fusion_comparison}, all fusion-based methods improve performance compared to the baseline without phenotypic fusion, confirming the benefit of incorporating individual-level information. Notably, the gated fusion approach achieves the highest overall accuracy, with a 6.3\% improvement over the non-fusion baseline. It also yields marked gains in AUC and specificity, suggesting enhanced discriminative ability and reduced false positives. Compared to standard fusion strategies, the gated mechanism offers consistent and meaningful improvements across all evaluation metrics.

These results demonstrate that gating provides a more effective and flexible means of integrating heterogeneous phenotypic features. By dynamically adjusting the contribution of each input, the model is better equipped to capture individual-specific variations relevant to the classification task.

\begin{table}[h]
\scriptsize
\centering
\caption{Performance of various phenotype feature fusion methods, including attention-based, additive, and gated fusion approaches.}
\setlength{\tabcolsep}{1.9mm}{
\begin{tabular}{lcccc}
\toprule
\textbf{Approach} & \textbf{ACC (\%, $\uparrow$)} & \textbf{AUC (\%, $\uparrow$)} & \textbf{SEP (\%, $\uparrow$)} & \textbf{SEN (\%, $\uparrow$)} \\
\midrule
No Feature Fusion & 77.1 $\pm$ 0.4 & 76.7 $\pm$ 0.4 & 82.1 $\pm$ 1.4 & 71.3 $\pm$ 1.1 \\
\midrule
Attention Fusion & 82.8 $\pm$ 0.6 & 82.5 $\pm$ 0.6 & 86.3 $\pm$ 1.7 & 78.7 $\pm$ 1.7 \\
Add Fusion & 82.5 $\pm$ 0.7 & 82.2 $\pm$ 0.8 & 86.3 $\pm$ 0.5 & 78.2 $\pm$ 1.4 \\
Concat Fusion & 83.0 $\pm$ 0.7 & 82.7 $\pm$ 0.7 & 87.3 $\pm$ 1.5 & 78.1 $\pm$ 1.4 \\
Gated Fusion (Ours) & \textbf{83.4 $\pm$ 0.4} & \textbf{83.0 $\pm$ 0.3} & \textbf{88.9 $\pm$ 0.5} & \textbf{77.1 $\pm$ 0.1} \\
\bottomrule
\end{tabular}
}
\label{tab:fusion_comparison}
\end{table}

{
\subsubsection{Ablation Study on the Use of Phenotypic Data}

To evaluate the contribution of phenotypic information, we conducted an ablation experiment where only neuroimaging data were used in the second stage. In this setting, population graph edges were constructed solely based on the similarity of learned brain features, and all phenotypic variables were excluded from the fusion process.
}

\begin{table}[h]
\scriptsize
\centering

\caption{Performance comparison with and without phenotypic data.}
\setlength{\tabcolsep}{1.6mm}{
\begin{tabular}{lcccc}
\toprule
\textbf{Method} & \textbf{ACC (\%, $\uparrow$)} & \textbf{AUC (\%, $\uparrow$)} & \textbf{SPE (\%, $\uparrow$)} & \textbf{SEN (\%, $\uparrow$)} \\
\midrule
Without Phenotypic Data & $76.7 \pm 0.7\%$ & $76.5 \pm 0.6\%$ & $73.4 \pm 2.2\%$ & $79.5 \pm 2.5\%$ \\
With Phenotypic Data & $\mathbf{83.4 \pm 0.4\%}$ & $\mathbf{83.0 \pm 0.3\%}$ & $\mathbf{77.1 \pm 0.1\%}$ & $\mathbf{88.9 \pm 0.5\%}$ \\
\bottomrule
\end{tabular}
}
\label{tab:without_phenotypic}

\end{table}

{
As shown in Table~\ref{tab:without_phenotypic}, even without phenotypic information, the two-stage framework outperforms the independently trained brain feature model. This observation is consistent with previous findings suggesting that imaging data alone provide substantial predictive power for disease classification \citep{shi2022nonimage}. Nevertheless, incorporating non-imaging variables further enhances the model’s capacity to capture inter-individual variability and improves overall classification performance.
}

\subsection{Analysis of Loss Function in Two Stage}

To investigate the impact of incorporating similarity loss in different training stages, we compare the performance of the model with and without this loss in both Stage 1 and Stage 2. The similarity loss is designed to enforce consistency in feature representations between similar instances, potentially enhancing the discriminative power of learned embeddings.

Table~\ref{tab:similarity_loss} summarizes the classification performance under four configurations: with and without similarity loss in Stage 1, and with and without similarity loss in Stage 2. As shown in the table, adding similarity loss in Stage 1 leads to only a marginal improvement in accuracy. This suggests that the similarity loss alone has a limited impact when used in the early stage of feature learning, where features are still coarse and less structured. In contrast, the addition of similarity loss in Stage 2, where graph construction is based on feature similarity, results in a more significant performance gain. Specifically, Stage 2 with similarity loss improves accuracy and AUC by 1.4\% compared to its counterpart without similarity loss. This indicates that the similarity constraint becomes more effective when the model already leverages similarity-based structures, reinforcing the consistency of neighboring nodes in the population-level graph. Overall, these findings demonstrate that the benefit of similarity loss is more pronounced in the second stage, where feature similarity directly contributes to the graph structure and phenotypic fusion.

\begin{table}[ht]
\centering
\scriptsize
\caption{Performance comparison under four training configurations: with or without similarity loss in Stage 1 and Stage 2.}
\setlength{\tabcolsep}{1.9mm}{
\begin{tabular}{lcccc}
\toprule
\textbf{Configuration} & \textbf{ACC(\%, $\uparrow$)} & \textbf{AUC(\%, $\uparrow$)} & \textbf{SPE (\%, $\uparrow$)} & \textbf{SEN (\%, $\uparrow$)} \\
\midrule
Stage 1 w/o similarity & 73.4 $\pm$ 0.9 & 72.8 $\pm$ 1.0 & 81.3 $\pm$ 1.6 & 64.3 $\pm$ 2.8 \\
Stage 1 w/ similarity  & 73.7 $\pm$ 0.6 & 73.2 $\pm$ 0.6 & 80.6 $\pm$ 1.8 & 65.9 $\pm$ 2.1 \\
Stage 2 w/o similarity & 82.0 $\pm$ 0.6 & 81.6 $\pm$ 0.6 & 86.6 $\pm$ 1.2 & 76.6 $\pm$ 1.5 \\
Stage 2 w/ similarity  & \textbf{83.4 $\pm$ 0.4} & \textbf{83.0 $\pm$ 0.3} & \textbf{88.9 $\pm$ 0.5} & \textbf{77.1 $\pm$ 0.1} \\
\bottomrule
\end{tabular}
}
\label{tab:similarity_loss}
\end{table}

\subsection{Interpretive Analysis}
In this subsection, we analyzed the interpretability of our proposed model in identifying critical nodes and node reassignment groups for brain graphs. 

\subsubsection{Node Importance}

To identify the brain regions most indicative of each disorder, we applied Grad-CAM \citep{Selvaraju17} to compute the gradient of the model's output with respect to node features after the first GAT layer. The top-10 most influential nodes for ASD, ADHD, and MDD are visualized in Figure~\ref{fig:important node}, with detailed region names listed in Table~\ref{tab:top10_regions}. These regions were further analyzed to interpret the disorder-specific and shared neurobiological patterns learned by the model.

\begin{table*}[ht]
\centering
\scriptsize
\caption{Top‑10 most predictive brain regions in ASD, ADHD, and MDD classification models.}
\vspace{10pt}
\begin{tabular}{@{}p{3.5cm} p{4.5cm} p{4.5cm} p{4.5cm}@{}}
\toprule
\textbf{Functional Network} 
  & \textbf{ASD Regions} 
  & \textbf{ADHD Regions} 
  & \textbf{MDD Regions} \\ 
\midrule
\textbf{Frontoparietal Network} 
  & -- 
  & --
  & Left Middle Frontal Gyrus (MFG.L), Right Inferior Frontal Triangular (IFGtriang.R), Left Inferior Frontal Triangular (IFGtriang.L), Left Middle Orbital Gyrus (ORBmid.L) \\
\addlinespace

\textbf{Default Mode Network} 
  & Right Frontal Pole (FP.R) 
  & Right Frontal Pole (FP.R) 
  & Right Angular Gyrus (ANG.R), Left Angular Gyrus (ANG.L), Right Precuneus (PCUN.R) \\
\addlinespace

\textbf{Limbic Network} 
  & Right Planum Polare (PP.R), Left Planum Polare (PP.L), Right Accumbens (Accbns.R), Left Accumbens (Accbns.L) 
  & Right Subcallosal Cortex (SC.R), Right Accumbens (Accbns.R), Left Accumbens (Accbns.L) 
  & Right Superior Orbital Gyrus (ORBsup.R) \\
\addlinespace

\textbf{Visual Network} 
  & Right Occipital Pole (OP.R) 
  & Right Occipital Pole (OP.R) 
  & Right Superior Occipital Gyrus (SOG.R) \\
\addlinespace

\textbf{Basal Ganglia Network} 
  & Right Caudate (Caud.R), Left Pallidum (Pall.L), Right Pallidum (Pall.R), Left Putamen (Put.L) 
  & Right Caudate (Caud.R), Left Caudate (Caud.L), Left Pallidum (Pall.L), Right Thalamus (Thal.R), Left Thalamus (Thal.L) 
  & -- \\
\addlinespace

\textbf{Sensorimotor Network} 
  & --
  & --
  & Right Precentral Gyrus (PreCG.R) \\
\bottomrule
\end{tabular}
\label{tab:top10_regions}
\end{table*}

\begin{figure}[ht]
    \centering
    \includegraphics[width=0.85\linewidth]{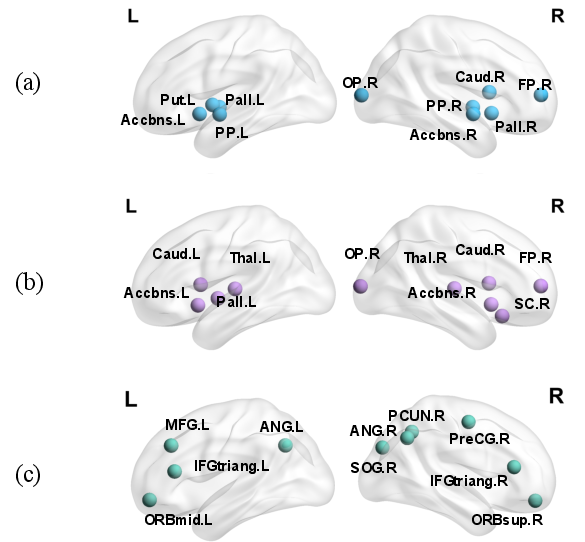} 
     \caption{Visualization of the top-10 most frequently identified brain region pairs across disorders, shown in both right and left hemisphere views. (a) In the ASD model, the most important regions are located within the limbic and basal ganglia networks. (b) In the ADHD model, key regions are similarly concentrated in the limbic and basal ganglia networks. (c) In the MDD model, the most influential regions are primarily distributed within the frontoparietal and default mode networks.}
    \label{fig:important node}
\end{figure}

For the two neurodevelopmental disorders, ASD and ADHD, the most predictive brain regions are predominantly located within the limbic and basal ganglia networks. Among these, the nucleus accumbens (Accbns.R, Accbns.L), a key region involved in reward processing and motivation, was consistently identified across both models. This structure has been implicated in social motivation deficits in ASD and in impulsive decision-making in ADHD \citep{dichter2012reward, dias2013reward}. Similarly, the caudate nucleus (Caud.R, Caud.L), which plays a critical role in motor control and executive functioning, also emerged as a key region. Prior studies have reported aberrant activation patterns of the caudate in both ASD and ADHD \citep{o2016decreased}, aligning with our findings. The frontal pole (FP.R), part of the default mode network (DMN), was also identified as a shared high-impact region across both disorders. This region is involved in higher-order cognitive and affective regulation and is frequently altered in neurodevelopmental conditions \citep{hrvoj2019neurodevelopmental}. In addition, the occipital pole (OP.R), belonging to the visual network, was identified as important in both ASD and ADHD, reflecting known abnormalities in visual processing \citep{jung2019decreased, wu2020role}.

Despite these commonalities, several disorder-specific patterns were observed. In ASD, the planum polare (PP.R, PP.L), a region involved in auditory and language processing, was ranked highly. This finding is consistent with prior reports of atypical auditory function in individuals with autism \citep{abrams2019impaired}. In contrast, the thalamus (Thal.R, Thal.L) was more prominent in the ADHD model. As a central hub for relaying sensory and motor signals, thalamic dysfunction has been associated with attentional deficits and treatment responsiveness in ADHD \citep{bailey2015role}. These findings suggest that while ASD and ADHD share alterations in core subcortical circuits, each disorder also involves distinct functional substrates.

In comparison, the most discriminative brain regions for MDD were primarily located within the frontoparietal network (FPN) and the default mode network (DMN), two canonical resting-state networks consistently implicated in mood disorders. The FPN is involved in high-level cognitive functions, including working memory and cognitive control, while the DMN supports self-referential thought and affective processing \citep{belleau2023default}. Dysregulation within and between these networks has been widely reported in MDD and is thought to contribute to core symptoms such as rumination and cognitive inflexibility \citep{marchetti2012default}.

In summary, our model identifies both shared and disorder-specific neurofunctional signatures. ASD and ADHD show substantial overlap, particularly in the limbic and basal ganglia networks, while MDD is primarily associated with disruptions in large-scale cognitive and affective systems. These findings underscore both convergent and divergent neurobiological pathways across diagnostic categories and suggest potential network-level targets for biomarker discovery and interventions.

\subsubsection{Group Reassignment}

To identify interpretable groupings, we calculated the probability of each node pair being assigned to the same group across samples and selected the top 10 pairs with the highest co-assignment frequency. These frequently co-assigned pairs are visualized in Figure~\ref{fig:enter-label}. To further examine the underlying structure, we organized these pairs into connected components, which reflect clusters of brain regions with similar feature representations and a high likelihood of joint assignment. The details of these components are provided in Table~\ref{tab:groups_all_disorders}.

\begin{figure}
    \centering
    \includegraphics[width=1\linewidth]{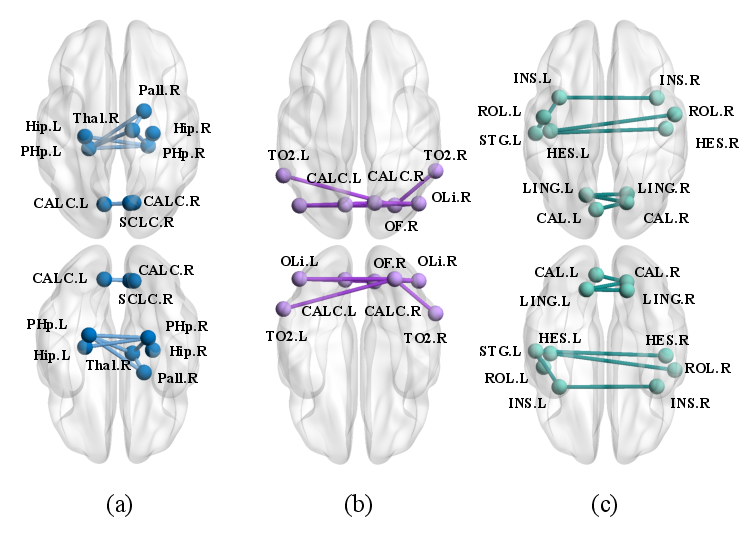}
    \caption{Brain region connections based on the top-10 highest-frequency node pairs, visualized in dorsal and ventral view. (a) The ASD model demonstrates reassigned connections primarily located in subcortical regions. (b) The ADHD model shows reassignments concentrated in the occipital lobe. (c) The MDD model exhibits grouping in prefrontal regions.}
    \label{fig:enter-label}
\end{figure}

\begin{table*}[h]
\centering
\scriptsize
\renewcommand{\arraystretch}{1.5}
\caption{Top-10 high-frequency node pair groups across ASD, ADHD, and MDD models.}
\vspace{5pt}
\begin{tabular}{@{}p{1.2cm} p{0.8cm} p{3cm} p{12cm}@{}}
\toprule
\textbf{Disorder} & \textbf{Group} & \textbf{Functional Network} & \textbf{Brain Regions} \\
\midrule
\multirow{4}{*}{ASD} 
  & 1 & Visual Network & Right Intracalcarine Cortex (Calc.R), Left Intracalcarine Cortex (Calc.L), Right Supracalcarine Cortex (SCLC.R) \\
  \cline{2-4}
  & 2 & Basal Ganglia Network & Right Thalamus (Thal.R), Right Pallidum (Pall.R) \\
  &   & Limbic Network & Left Hippocampus (Hip.L), Right Hippocampus (Hip.R) \\
  &   & Default Mode Network & Left Parahippocampal Gyrus (PHG.L), Right Parahippocampal Gyrus (PHG.R) \\
\cline{1-4}
\multirow{2}{*}{ADHD} 
  & \multirow{2}{*}{1} & \multirow{2}{*}{Visual Network} & Right Intracalcarine Cortex (Calc.R), Left Intracalcarine Cortex (Calc.L), Right Lingual Gyrus (Ling.R), Left Lingual Gyrus (Ling.L), Right Cuneal Cortex (Cun.R), Left Cuneal Cortex (Cun.L), Right Supracalcarine Cortex (SCLC.R) \\
\cline{1-4}
\multirow{5}{*}{MDD} 
  & 1 & Sensorimotor Network & Left Rolandic Operculum (ROL.L), Right Rolandic Operculum (ROL.R), Left Heschl Gyrus (HES.L), Right Heschl Gyrus (HES.R), Left Superior Temporal Gyrus (STG.L) \\
  &   & Ventral Attention Network & Left Insula (INS.L), Right Insula (INS.R) \\
  \cline{2-4}
  & 2 & Visual Network & Left Calcarine Cortex (CAL.L), Right Calcarine Cortex (CAL.R), Left Lingual Gyrus (LING.L), Right Lingual Gyrus (LING.R) \\
\bottomrule
\end{tabular}
\label{tab:groups_all_disorders}
\end{table*}

The ASD model revealed frequent co-assignment of regions spanning the basal ganglia, limbic, visual, and default mode networks. Notably, subcortical structures such as the pallidum (Pall.L, Pall.R), thalamus (Thal.R), and hippocampus (Hip.L, Hip.R) emerged as highly connected, indicating disruptions in motor coordination, memory, and socio-emotional processing \citep{pessoa2017network}. The ADHD model showed a dominant group localized within the visual network. Frequently co-assigned regions included the bilateral intracalcarine cortex (Calc.L, Calc.R), lingual gyrus (Ling.L, Ling.R), cuneal cortex (Cun.L, Cun.R), and supracalcarine cortex (SCLC.R). These anatomically adjacent occipital regions are functionally linked to early-stage visual processing \citep{GAGE201817}. For MDD, the principal grouping encompassed regions from the sensorimotor and ventral attention networks. Frequently co-assigned areas included the Rolandic operculum (ROL.L, ROL.R), Heschl’s gyrus (HES.L, HES.R), superior temporal gyrus (STG.L), and bilateral insula (INS.L, INS.R), forming an integrated network that mediates sensorimotor processing, auditory integration, and affective salience detection \citep{dziedzic2022anatomical, have2007heschl}.

These region groupings emerged from the model’s learned co-assignment patterns, suggesting that they may share similar feature representations or jointly aid in distinguishing disease.

\section{Discussion and Conclusion}

In this study, we proposed a B2P-GL framework to enhance brain network representation and reduce individual bias in brain disorder analysis using medical imaging. In the first stage, termed brain representation learning, we leveraged the knowledge embedded in brain atlases, refining the graph through node reassignment to improve brain representation. In the second stage, termed population disorder diagnosis, we incorporated personal characteristics to estimate confounding effects and fused these with brain features. Our framework outperformed various GNNs in FC-based disorder identification and provided clinical interpretability, demonstrating significant potential for application in the analysis of diverse neurodevelopmental disorders.

Overall, our work demonstrates a notable level of superiority. (1) Experiments on three publicly available brain disorder datasets show that our framework consistently outperforms existing GNN-based methods in functional connectivity-based classification, confirming its robustness and broad applicability. (2) Ablation studies further highlight the critical roles of our key design components, including semantic brain graph refinement and node reassignment in the first stage, which enhance the expressiveness of brain representations, as well as condition-based population modeling for confound mitigation and gated fusion for integrating multi-modal phenotypic data in the second stage. (3) The interpretive analysis reveals meaningful patterns in brain functional connectivity, offering valuable insights into disorder-specific alterations and enhancing the clinical relevance of the learned representations.

Despite these promising findings, several limitations remain. The brain region embeddings used in our method are derived from general-purpose large language models, which may not fully reflect domain-specific anatomical and functional nuances. Incorporating structured neuroscience knowledge, such as expert-curated ontologies or dedicated anatomical corpora, could yield more robust and biologically grounded representations. Additionally, the variability inherent in large language models introduces uncertainty that requires further control and systematic validation. Future work could explore domain-adaptive fine-tuning strategies, and establish comprehensive protocols to ensure the factual consistency of generated descriptions. It would also be valuable to compare different atlases, including anatomical and multi-atlas schemes, and to extend the framework to other imaging modalities or clinical conditions for a more comprehensive evaluation of its general utility. In addition, integrating advanced GNN architectures, such as contrastive and self-supervised learning, could enhance the capacity to model complex relationships and improve generalizability.


\section*{CRediT authorship contribution statement}
\textbf{Qianqian Liao}: Conceptualization, Methodology, Formal analysis, Visualization, Writing-Original draft, Writing-Review $\&$ Editing. \textbf{Wuque Cai}: Writing-Original draft, Writing-Review $\&$ Editing, Methodology. \textbf{Hongze Sun}: Visualization, Formal analysis, Writing-Review $\&$ Editing. \textbf{Dongze Liu}: Data Curation, Formal analysis. \textbf{Duo Chen}: Funding acquisition, Visualization. \textbf{Dezhong Yao}: Conceptualization, Funding acquisition, Supervision, Writing-Review $\&$ Editing. \textbf{Daqing Guo}: Conceptualization, Supervision, Funding acquisition, Resources, Writing-Original draft, Writing-Review $\&$ Editing.

\section*{Declaration of Competing Interest}
The authors declare that they have no known competing financial interests or personal relationships that could have appeared to influence the work reported in this paper.

\section*{Data and code availability}
We will share our code at https://github.com/GuoLab-UESTC after this manuscript is accepted for publication. All the datasets used in our work are sourced from public datasets.


\section*{Acknowledgments}

This work was supported in part by the STI 2030–Major Projects under grant 2022ZD0208500, in part by the National Key Research and Development Program of China under grant 2023YFF1204200, in part by the Sichuan Science and Technology Program under grant 2024NSFTD0032,  grant 2024NSFJQ0004 and grant DQ202410, and in part by the Natural Science Foundation of Chongqing, China under grant CSTB2024NSCQ-MSX0627, in part by the Science and Technology Research Program of Chongqing Education Commission of China under Grant KJZD-K202401603, and in part by the China Postdoctoral Science Foundation under grant 2024M763876.

\bibliographystyle{elsarticle-harv} 
\bibliography{ref.bib}

\end{document}